%% file: ecjsample.tex
\documentclass[twoside]{article}
\usepackage{ecj,palatino,epsfig,latexsym,natbib}
\usepackage[dvipsnames]{xcolor}
\usepackage{graphicx}
\usepackage{tikz}
\usepackage{xcolor}
\usepackage{amsmath}
\usepackage{amssymb}
\usepackage{bbm}
\usepackage{booktabs}
\usepackage{multirow}
\usepackage{colortbl}
\usepackage{adjustbox}
\usepackage{paralist}
\usepackage[normalem]{ulem}
\usepackage{hyperref}
\usepackage{subfig}

\newcommand{\added}[1]{\textcolor{black}{#1}}
\newcommand{\deleted}[1]{}

\renewcommand\vec{\mathbf}

\newcommand{\spaceELA}[0]{\mathcal{F}_\text{ELA}} 
\newcommand{\spaceDELA}[0]{\mathcal{F}_\text{D-ELA}} 
\newcommand{\spaceA}[0]{\mathcal{A}} 
\newcommand{\spaceF}[0]{\mathcal{F}} 
\newcommand{\spaceP}[0]{\mathcal{P}} 
\newcommand{\spaceI}[0]{\mathcal{I}} 
\newcommand{\spaceT}[0]{\mathcal{T}} 
\newcommand{\spaceX}[0]{\mathcal{X}} 
\newcommand{\spaceY}[0]{\mathcal{Y}} 

\newcommand{\nELA}[0]{{c_\text{ELA}}}
\newcommand{\nDELA}[0]{{c_\text{D-ELA}}}

\newcommand{\nA}[0]{{n_\mathcal{A}}}
\newcommand{\nF}[0]{{n_\mathcal{F}}}

\newcommand{\R}[0]{\mathbb{R}} 
\newcommand{\T}[0]{\text{T}}

\DeclareMathOperator*{\argmin}{\text{arg\,min}}

\begin{document}

\ecjHeader{x}{x}{xxx-xxx}{201X}{Pretrained Transformers for Deep Exploratory Landscape Analysis}{M.V. Seiler et al.}
\title{\bf Deep-ELA: Deep Exploratory Landscape Analysis with Self-Supervised Pretrained Transformers for Single- and Multi-Objective Continuous Optimization Problems}  

\author{\name{\bf Moritz Vinzent Seiler} \hfill \addr{moritz.seiler@uni-paderborn.de}\\ 
        \addr{\added{Machine Learning and Optimisation, Paderborn University, Germany}}
\AND
       \name{\bf Pascal Kerschke} \hfill \addr{pascal.kerschke@tu-dresden.de}\\
        \addr{Big Data Analytics in Transportation, TU Dresden, Germany;\\
        ScaDS.AI Dresden/Leipzig, Germany}
\AND
       \name{\bf Heike Trautmann} \hfill \addr{heike.trautmann@uni-paderborn.de}\\
        \addr{Machine Learning and Optimisation, Paderborn University, Germany; \\ Data Management and Biometrics Group, University of Twente, Netherlands}
}

\maketitle

\begin{abstract}

In many recent works, the potential of \textit{Exploratory Landscape Analysis}~(ELA) features to numerically characterize, in particular, single-objective continuous optimization problems has been demonstrated. These numerical features provide the input for all kinds of machine learning tasks on continuous optimization problems, ranging, i.a., from \textit{High-level Property Prediction} to \textit{Automated Algorithm Selection} and \textit{Automated Algorithm Configuration}. Without ELA features, analyzing and understanding the characteristics of single-objective continuous optimization problems \deleted{would be impossible} \added{is --- to the best of our knowledge ---  very limited}.  

Yet, despite their \deleted{undisputed} usefulness\added{, as demonstrated in several past works}, ELA features suffer from several drawbacks. These include, in particular, (1.) a strong correlation between multiple features, as well as (2.) its very limited applicability 
to multi-objective continuous optimization problems. 
As a remedy, recent works proposed deep learning-based approaches as alternatives to ELA. In these works, e.g., point-cloud transformers were used to characterize an optimization problem's fitness landscape. However, these approaches require a large amount of labeled training data.


Within this work, we propose a hybrid approach, \textit{Deep-ELA}, which combines (the benefits of) deep learning and ELA features. Specifically, we pre-trained four transformers on millions of randomly generated optimization problems to learn deep representations of the landscapes of continuous single- and multi-objective optimization problems. Our proposed framework can either be used out-of-the-box for analyzing single- and multi-objective continuous optimization problems, or subsequently fine-tuned to various tasks focussing on algorithm behavior and problem understanding.

\end{abstract}


\section{Introduction and Related Work}
Optimization problems, often found at the heart of numerous scientific and industrial applications, present challenges that necessitate robust and efficient problem-solving methodologies. A central concern in this field is the analysis and characterization of \deleted{the continuous} optimization problem\added{s'} landscapes\added{, i.e. continuous single objective problems but also i.a. combinatorial optimization, mixed-integer or multi-objective optimization}. One methodological concept that has emerged as particularly significant for such a characterization is \textit{Exploratory Landscape Analysis}~\citep[ELA;][]{mersmann2011exploratory}. ELA facilitates a numerical representation of single-objective continuous optimization problems, providing insights into their intrinsic characteristics. The numerical features harvested from ELA serve as the foundation for various machine-learning tasks pertinent to continuous optimization. Such tasks include, but are not limited to, \textit{High-level Property Prediction}~\citep[HL2P;][]{mersmann2011exploratory,seiler2022collection,volz2023tools}, \textit{Automated Algorithm Selection}~\citep[AAS;][]{Rice1976AS,kerschke2019automated}, and 
\textit{Automated Algorithm Configuration}~\citep[AAC;][]{hutter2009paramils,huang2019survey,schede2022survey}.

While ELA has undoubtedly transformed how researchers approach single-objective continuous optimization problems, it is not without its challenges. Two of the most pronounced limitations include the strong correlations between numerous features and their restricted use in the multi-objective domain. The advent of deep learning has ushered in a myriad of solutions across disciplines. Recent studies have proposed leveraging deep learning, specifically point-cloud transformers, as a potential remedy to the shortcomings of ELA \citep{seiler2022collection,prager2022automated}. However, such approaches, while promising, come with their own set of challenges, most notably the requirement of vast amounts of labeled training data.

Given the strengths and weaknesses of both, ELA and deep learning-based methodologies, there is a compelling case to be made for a synthesis of the two. This paper seeks to bridge this gap by introducing a hybrid approach, capitalizing on the merits of both paradigms. We thus present a novel method that employs four pre-trained transformer models tailored to extract deep representations of the landscapes of both single- and multi-objective continuous optimization problems. This innovative approach promises enhanced flexibility and efficacy \deleted{in the realm} of optimization problem analysis and understanding \added{as these models can be used out-of-the-box without the need for additional feature selection or normalization. In addition, learned features do not require expertise to design meaningful feature sets. Instead, one can simply fine-tune Deep ELA on novel optimization problems that are not meaningfully represented by classical ELA }.

We introduce these pre-trained transformers as  \textit{Deep Exploratory Landscape Analysis}~(Deep-ELA).\footnote{\added{An extension of Deep-ELA to the \texttt{pflacco} package can be found here: \url{https://github.com/mvseiler/deep\_ela.git}.}} The strengths of Deep-ELA are prominently seen in its inherent adaptability to multi-objective continuous optimization problems and the reduced correlation among its learned features. We demonstrate the straightforward adaptability of Deep-ELA by applying it to rigorous tests across three distinct case studies that showcase their efficacy:

\begin{compactenum}
    \item A classification task involving problem instances from the \textit{Black-box Optimization Benchmarking Suite}~\citep[BBOB;][]{Hansen2009bbob}. Here, we focus on modeling and predicting their \textit{High-Level Properties}, effectively revisiting the experiments previously conducted by \citet{mersmann2011exploratory} and \citet{seiler2022collection}.
    \item An assessment of single-objective AAS on BBOB, featuring twelve algorithms sourced from the \textit{COmparing Continuous Optimisers}~\citep[COCO;][]{nikolaus2019coco} platform. This particular study echoes the works of \citet{kerschke2019comprehensive} and later \citet{prager2022automated}.
    \item An AAS study, applied to a set of multi-objective optimization problems, utilizing seven distinct algorithms. While this study draws inspiration from \citet{rook2022potential}, it ventures into a different direction, focusing on AAS rather than AAC.
\end{compactenum}

\noindent The remainder of this paper is structured as follows. Initially, we equip readers with a concise background on ELA, providing an overview of feature-free alternatives and laying out the notation employed throughout the paper (Section~\ref{sec:BG}). Subsequently, Section~\ref{sec:DeepELA} dives into the architectural essence of Deep-ELA. Following this, we delve into a thorough exploration of the various datasets featured in this work in Section~\ref{sec:Data}. We then pivot to a comparative analysis, contrasting the outcomes of Deep-ELA with traditional ELA across the previously mentioned case studies in Section~\ref{sec:exp}. In Section~\ref{sec:Disc}, we present our conclusions and provide opportunities for future research.

\section{Background}\label{sec:BG}
In the following, we will briefly outline the notation of this paper. An optimization function\added{, $Z$,} is given by:
\begin{align}
    Z: \spaceX \to \spaceY, \quad \vec x \mapsto (z_1(\vec x), z_2(\vec x), \ldots,  z_m(\vec x))^\top
\end{align}
where $\spaceX \subseteq \mathbb{R}^d$ is the decision and $\spaceY \subseteq \mathbb{R}^m$ the objective space. Every scalar function $z_i: \spaceX \to \R$, with $i\in\{1,\ldots,m\}$, symbolizes a distinct objective function that maps from the decision space $\spaceX$ to a real value. In this context, $\vec x \in \spaceX$ is referred to as a \textit{candidate solution}, and the set $X = \{\vec x_1,\vec x_2, \ldots, \vec x_n\} \subseteq \mathcal{X}$ is formed of $n$ such solutions. An objective or fitness vector is represented by $\vec y = Z(\vec x)$ with $\vec y \in \spaceY$ and a length of $m$. If $m>1$, then $Z$ is a multi-objective function, whereas for $m = 1$, $\vec y$ is a scalar, and $Z$ is a single-objective function. Finally, $Y = \{Z(\vec x) \mid \vec x \in X\}$ 
is a set of objective values.

In single-objective optimization, the task is defined as:
\begin{align}
    \vec x^* = \argmin_{\vec x \in X} Z(\vec x)
\end{align}
Here, $\vec x^*$ is \deleted{the} \added{an} optimal solution for $Z$. The goal of single-objective optimization is to find a solution $\vec x^*$ that w.l.o.g. minimizes the function $Z$ globally. If multiple optimal solutions exist, $Z$ is \textit{multimodal}, and the aim shifts to locating all optimal solutions.

In contrast, in the multi-objective setting, optimizing for one objective can negatively affect another. Therefore, any solution provides a trade-off between the objectives, and the goal of multi-objective optimization is to find a set of optimal solutions, known as a non-dominated or Pareto set. Given two candidate solutions $\vec x_i$ and $\vec x_j$, Pareto dominance, denoted as $\vec x_i \prec \vec x_j$, means $\vec x_i$ dominates $\vec x_j$, if it is equal or better in all objectives, and strictly better in at least one of them. Formally, this is represented as:
\begin{align}
    z_k(\vec x_i) &\leq z_k(\vec x_j) \quad \forall k \in \{1, \ldots, m\}, \text{ and }\\
    z_l(\vec x_i) &< z_l(\vec x_j) \quad \exists\: l \in \{1, \ldots, m\}.
\end{align}
The goal of multi-objective optimization is to determine the Pareto set 
\begin{align}
    X_E = \{\vec x_i \in X \mid \nexists \vec x_j \in X: \vec x_j \prec \vec x_i\}
\end{align}
and its mapping to the objective space, called Pareto front 
\begin{align}
    Y_E = \{Z(\vec x) \mid \vec x \in X_E\}.
\end{align}

In many studies concerning \textit{algorithm behavior} and \textit{problem understanding}, $Z$ is treated as a \textit{blackbox} optimization problem, implying that \deleted{direct formal analyses cannot be conducted on $Z$} \added{there is no algebraic expression of $Z$ given. Hence, no formal algebraic analysis can be conducted directly on $Z$}. Instead, preliminary analyses are often carried out on a small sample of candidate solutions 
$X$, which 
is usually derived using a random or quasi-random generator like Latin Hypercube Sampling \citep[LHS;][]{mckay1979comparison} or Sobol sequences \citep{sobol1967distribution}. 
The corresponding image $Y$ is obtained by evaluating the optimization problem $Z$ at the candidate solutions $X$. The tuple $(X,Y)$ is then available for subsequent studies.

Such studies span \textit{problem-} and \textit{algorithm behavior-understanding}, \textit{algorithm design}, and AAS. Hereafter, we will refer to these types of studies as \textit{downstream tasks} or \textit{downstream studies}.  This terminology is borrowed from the \textit{self-supervised} training community, which differentiates between two phases: \textit{(1.)} \textit{pre-training} and \textit{(2.)} \textit{fine-tuning}. Self-supervised learning addresses tasks with insufficient training datasets. Models are thus pre-trained using an auxiliary training task -- potentially using a synthetic set if the primary training set is inappropriate (e.g., too small or risk of overfitting). Subsequently, these models are applied or fine-tuned for the original \textit{downstream task}.

\subsection{Exploratory Landscape Analysis}
\begin{figure}[tb]
    \begin{minipage}[t]{0.49\textwidth}
        \input{figures/aas_classic}
    \end{minipage}
    \begin{minipage}[t]{0.49\textwidth}
        \input{figures/aas_novel}
    \end{minipage}
    \caption{Comparison of the feature-based (left) versus feature-free (right) approach on one common downstream task: \textit{Automated Algorithm Selection}~(AAS).
    In the realm of AAS, there is no single universally superior algorithm for all problem instances. Instead, AAS uses a portfolio of algorithms $\spaceA = \{A_1, \ldots, A_\nA \}$. The optimal selector is formally defined as $S\colon\spaceI\rightarrow\spaceA$. 
    Typically, the selector $S$ is trained using machine learning to optimize a given performance metric. However, standard machine learners in AAS cannot process raw problem instances directly, necessitating a transformation into numerical vectors. This transformation is given as $F: \spaceI \to \spaceF \subseteq \R^{\nF}$, where $F$ is a \textit{mapper} converting an instance $I\in\spaceI$ into a real-valued vector, termed \textit{(instance) features}, in the \textit{feature space} $\spaceF$. Therefore, in a standard AAS scenario, the selector is defined as $\dot S\colon\spaceF\rightarrow\spaceA$, accepting features rather than actual instances.}
    \label{fig:aas_comparison}
\end{figure}
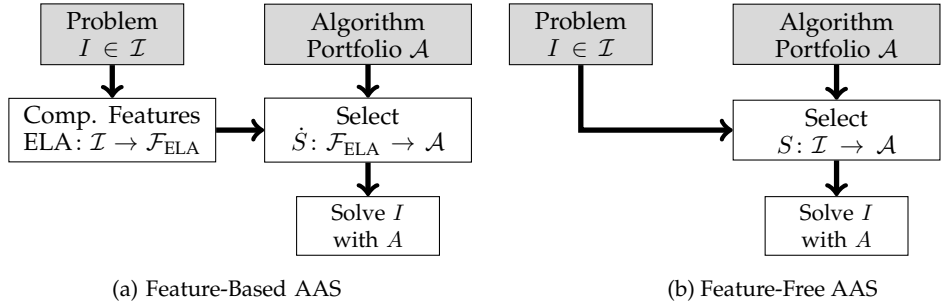
For downstream tasks involving continuous optimization problems, traditional machine learning methods aim to learn a specific underlying task. In AAS, for instance, classical machine learners like \textit{Support Vector Machines}~\citep[SVM;][]{cortes1995support}, \textit{Random Forests}~\citep[RF;][]{breiman2001random}, or \textit{$k$-Nearest Neighbors}~\citep[$k$NN;][]{cover1967nearest}, are utilized to address the \textit{Algorithm Selection Problem}~(ASP) (see Figure~\ref{fig:aas_comparison}a). \added{The difference between AAS and ASP is that while the latter refers to the challenge of choosing the best algorithm from a set of algorithms for a specific instance, based on performance metrics like speed or solution quality; the former addresses the use of machine learning to automatically select the best algorithm, typically based on problem features.} 
However, these classical learners cannot directly process raw samples of candidate solutions, i.e., $(X,Y)$, as the learners necessitate feature vectors as input per problem instance. Yet, transforming the set of (randomly distributed) points into a vector is not suitable, as such a mapping either (a) lacks an unambiguous order of the points, or (b) requires the points to be aligned in a grid structure, which in turn is affected by the \emph{curse of dimensionality}. 
Therefore, it is essential to convert the set of candidate solutions $(X,Y)$ into a vector of numerical values that provides a meaningful representation of the candidate tuple. This conversion, or transformation, is performed by a mapping function, or short \textit{mapper}, $F: (\spaceX,\spaceY) \to \spaceF \subseteq \R^{\nF}$, which translates the decision and objective values into a fixed-length real-valued vector, termed \textit{(instance) features}, in the \textit{feature space} $\spaceF$.

For continuous optimization problems, \textit{Exploratory Landscape Analysis}~\citep[ELA;][]{mersmann2011exploratory} features are commonly employed in many downstream tasks, including AAS. \added{ELA was termed by \citet{mersmann2011exploratory} but some feature sets that nowadays are considered as subsets of ELA were already introduced before, i.e. \textit{Dispersion}~\citep{lunacek2006dispersion} and \textit{Fitness Distance Correlation}~\citep{jones1995fitness}.} Formally, ELA is a function that maps samples $(X,Y)$ from the decision and objective space $(\spaceX, \spaceY)$ to the $\nELA$-dimensional feature space $\spaceELA \subseteq \R^{\nELA}$ 
\begin{align*}
    \text{ELA} \colon (\spaceX, \spaceY) \rightarrow \spaceELA.
\end{align*}
Over time, a plethora of ELA features for single-objective optimization problems emerged \citep{kerschke2023exploratory}, such as:
\begin{figure}
    \begin{minipage}[t]{0.545\textwidth}
        \centering
        \includegraphics[width=\textwidth, trim={18mm 40.5mm 22mm 0mm}, clip]{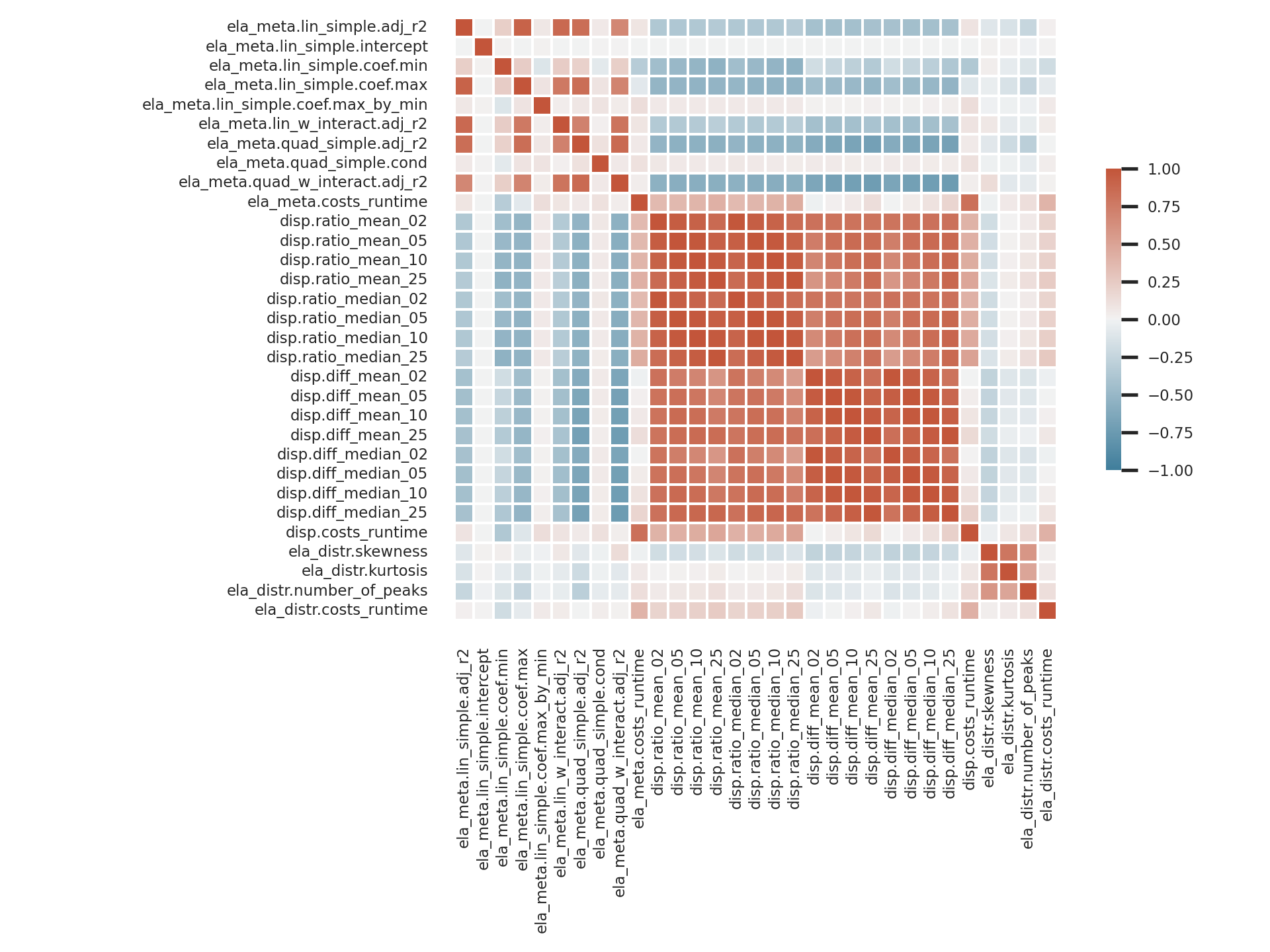}\\
        (a) ELA Features
    \end{minipage}
    \begin{minipage}[t]{0.45\textwidth}
        \centering
        \includegraphics[width=\textwidth, trim={14mm 10mm 10mm 2mm}, clip]{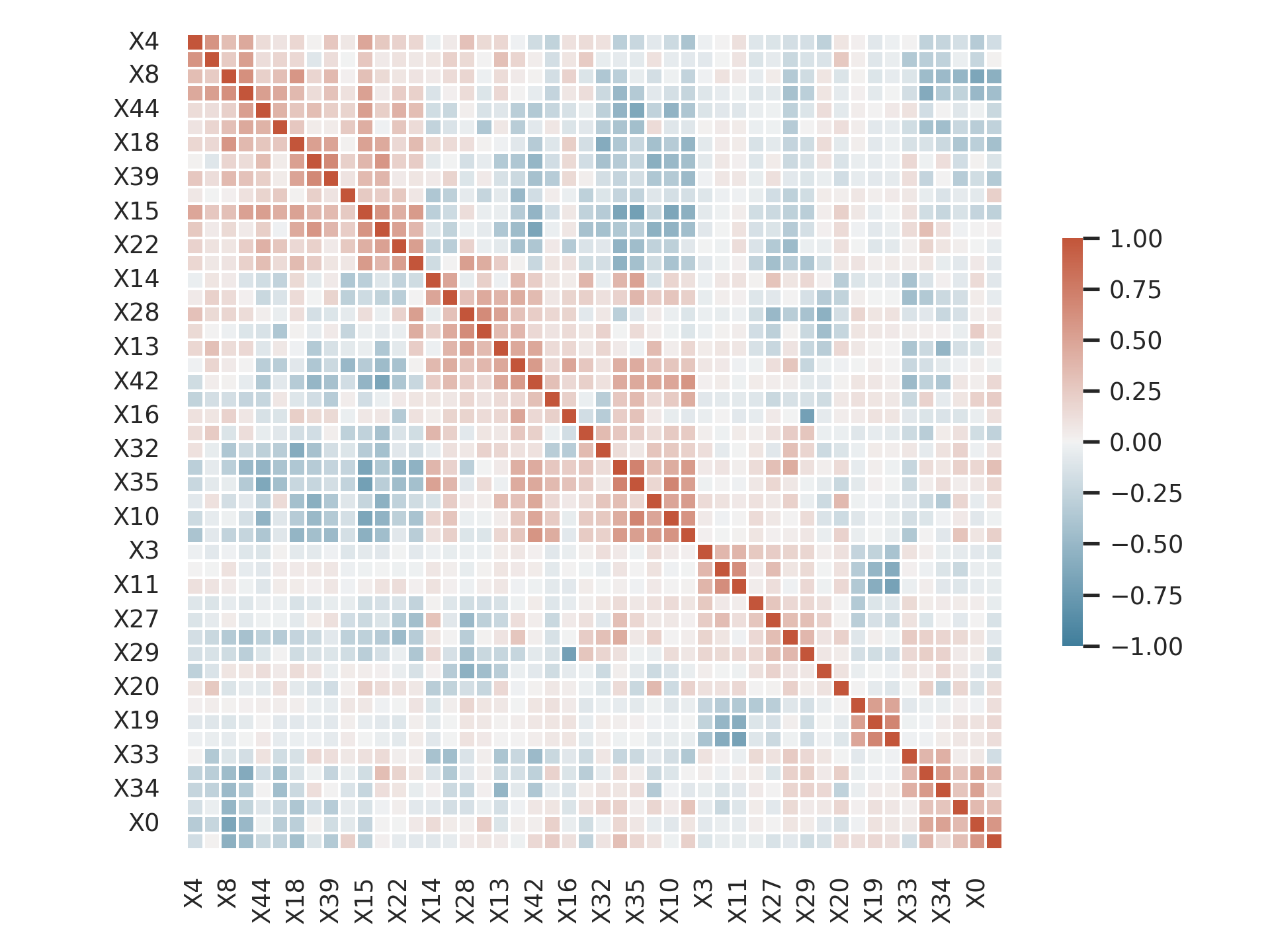}\\
        (b) Deep-ELA Features
    \end{minipage}

    \caption{Comparison of exemplary correlation matrices of (Deep-)ELA features on BBOB: (a) classical ELA features (here: \textit{meta-model}, \textit{dispersion} and \textit{$y$-distribution}), and (b) Deep-ELA features (Large-$50d$). Correlations are calculated across all 24 functions of the BBOB suite, but individually for every instance $1$ to $20$ and dimension $2,3,5,10$. Afterward, the 80 correlation maps are \texttt{mean}-aggregated.}
    \label{fig:corr_ela_vs_dela}
\end{figure}

\begin{compactenum}
    \item \textbf{Classical ELA} \citep{mersmann2011exploratory} is a superset of six different groups of ELA features, of which the \textit{meta-model}- and \textit{y-distribution}-features are the most commonly used ones.
    \item \textbf{Dispersion} \citep{lunacek2006dispersion} measures the spread in decision space of the best fraction of points in relation to the spread of all points of the full sample. 
    \item \textbf{Information Content} \citep{munoz2014exploratory} measures, i.a., the landscape's smoothness and ruggedness based on statistics summarizing a series of random walks.
    \item \textbf{Fitness Distance Correlation} \citep{jones1995fitness} measures the correlation between distances in the objective and distances in the decision space.
    \item \textbf{Nearest Better Clustering} \citep{kerschke2015detecting} compares the relation between the set of nearest neighbors in the decision space to the set of nearest better neighbors, where `better' relates to neighbors with lower objective values.
    \item \textbf{Miscellaneous} \citep{kerschke2019comprehensive} is a collection of different feature types such as features based on \textit{Principal Component Analysis}~(PCA).
\end{compactenum}

\begin{figure}[tb]
    \centering
    \includegraphics[width=0.65\textwidth, trim={0mm 0mm 0mm 0mm}, clip]{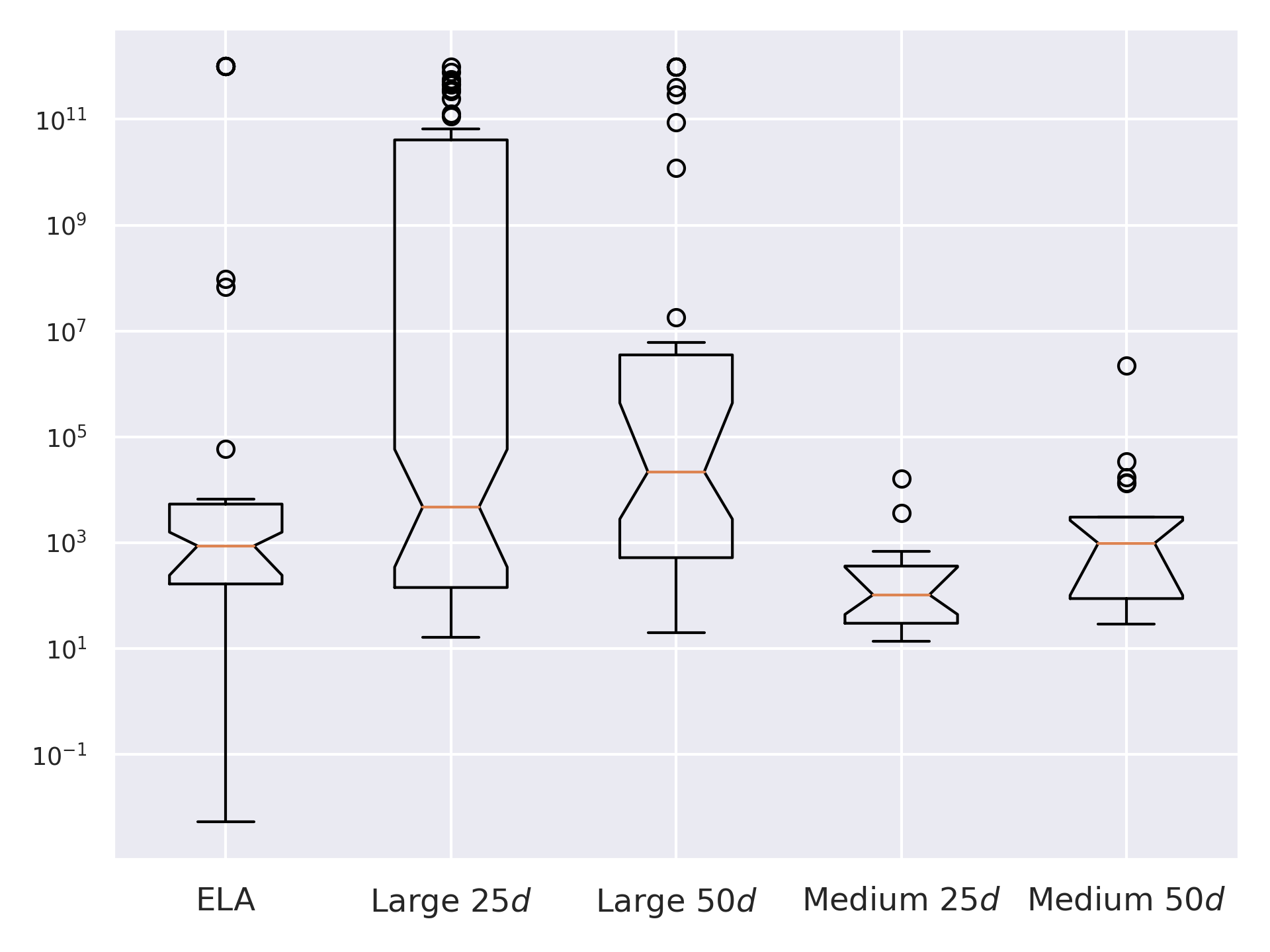}
    \caption{\textit{Signal to Noise Ratio}~(SNR) of ELA features (left boxplot) on BBOB compared to the SNR values of features from four Deep-ELA models. 
    We used $\text{SNR} = \mu^2 / \sigma^2$ with mean $\mu$ and standard deviation $\sigma$. 
    For $\sigma\simeq0$, values are imputed with $10^{12}$, which is the highest observed value. Higher values indicate lower noise.
    SNR values are calculated per feature based on instances $1$ to $20$ and then mean-aggregated over the $24$ functions and four dimensions ($2,3,5,10$). \added{Notches show the 95\% confidence intervals around the median. The large models yield the highest SNR while the medium models yield the lowest which is to be expected as the large models contain more parameters to create more sophisticated features. Classical ELA features are somewhat `in-between' while simultaneously containing features with multiple, very low SNR values.}
    }
    \label{fig:boxplot_snr}
\end{figure}

\deleted{While the utility of ELA features is widely recognized,}\added{Though ELA features are frequently used,} they also have some limitations \citep{kerschke2023exploratory}. As, e.g., discussed by \citet{Renau2019} and \citet{eftimov2020linear}, many ELA features exhibit strong correlations to each other, as illustrated in Figure~\ref{fig:corr_ela_vs_dela}a, or have a low \textit{Signal to Noise Ratio}~(SNR), see Figure~\ref{fig:boxplot_snr}. Moreover, some features from sets like \textit{Classical ELA}, \textit{Information Content}, and \textit{PCA} are sensitive to random scaling and shifting \citep{Renau2020}; see \citet{prager2023nullifying} for an in-depth analysis of ELA feature sensitivity. These issues necessitate \textit{feature selection} for many downstream tasks to eliminate noise or redundancy. Finally, ELA features are primarily tailored for single-objective optimization problems, not multi-objective ones. In fact, \citet{kerschke2016flacco} used ELA features to characterize bi-objective problems; however, they only considered simple test problems from DTLZ \citep{deb2005} and ZDT \citep{zitzler2000}, and applied ELA to the single-objective components of these problems, neglecting any interaction between the objectives.

\added{Another use-case for ELA features is \textit{Instance Space Analyzis} \citep[ISA;][]{smith2014towards}. It can be used to analyze the space that is covered by all instances within a given dataset to identify (dis-)similarities of structural properties between instances. \citet{smith2023instance} propose a framework that utilizes ELA features for ISA. On the other hand, ISA can also be used to identify empty regions that are not covered by existing optimization problems. \citep{prager2023neural} proposed a method to create artificial benchmark problems that cover specific regions within the instance space. These problem instances are learned by neural networks.}



\subsection{Deep Learning-Based Approaches as Alternative to ELA}

In the \textit{evolutionary computation}~(EC) community, methods that avert features are often termed \textit{feature-free}. 
Yet, in the deep learning domain, this term can be misleading. In the latter, \textit{features} concern internally learned features of complex deep learning models, whereas in the EC domain, (ELA) features refer to summary statistics quantifying the intrinsic characteristics of the given optimization problem (see Figure~\ref{fig:aas_comparison}b for an illustration of feature-free AAS). 

In recent studies, efforts have been made to solve downstream tasks without depending on ELA features, thus, sidestepping their limitations. 
In the context of discrete optimization (which is beyond the scope of this paper), \citet{Alissa2019ASWithoutFeatureExtraction} employed a \textit{Long Short-Term Memory}~\citep[LSTM;][]{hochreiter1997long} for AAS on the $1d$-Bin Packing problem, without pre-computed features. Similarly, \citet{seiler2020deep} used \textit{Convolutional Neural Networks}~\citep[CNN;][]{lecun1995convolutional} for AAS on the \textit{Traveling Salesperson Problems} (TSP) without pre-computed instance features.

Within continuous optimization, \citet{prager2021towards} introduced a method applying CNNs to two-dimensional single-objective optimization problems. This was expanded by \citet{seiler2022collection}, who presented image- and point cloud-based techniques for high-dimensional, single-objective optimization problems. Notably, the \textit{Point Cloud Transformer}~\citep[PCT; originally proposed by][]{guo2021pct} delivered top-tier results without the need for ELA features. Subsequently, \citet{prager2022automated} employed PCT alongside traditional ELA-based methods for AAS on continuous, single-objective optimization tasks. Only ELA-based \textit{Multi-Layer Perceptrons}~(MLP) with a carefully designed loss function managed to surpass the performance of the PCTs.

Compared to deep learning-based feature-free techniques, ELA-based methods usually require significantly fewer training instances under the assumption that these smaller benchmark sets are nonetheless representative and comprehensive. 
This becomes particularly advantageous when optimization problems are scarce or crafting a training dataset is computationally expensive. While data augmentation can amplify the training set, deep learning methods may still generalize poorly in case of limited training data. Note that in this context, \textit{small} means a few hundred and \textit{large} implies tens or even hundreds of thousands of instances\added{; e.g. \citet{seiler2022collection} utilized $144\,000$ training examples while \citet{prager2022automated} only relied on 4\,800 training examples (both including ten repetitions per instance).}

\section{Deep Exploratory Landscape Analysis}\label{sec:DeepELA}

In the following, we aim to merge the advantages of both \textit{feature-based} and \textit{feature-free} analysis in the context of continuous optimization. We introduce a large, pre-trained deep learning model trained on millions of randomly generated continuous, single- and multi-objective optimization instances. The generator for these tasks draws inspiration from \citet{tian2020recommender} and \citet{van2023doe2vec}, but with adaptations to suit our requirements. 
Our model's general goal is the automated extraction of instance features from initial samples of the given optimization instance. We dub this methodology \textit{Deep Exploratory Landscape Analysis} (Deep-ELA). \added{\citet{seiler2024synergies} analyzed these learned features in detail and, especially, analyzed the complementarity and synergies between classical and Deep ELA.} 

\subsection{\added{Outline of Deep ELA}}
Deep-ELA is trained to be invariant against scaling, shifts, and rotations, yielding higher SNR values and minimally correlated features (see Figure~\ref{fig:corr_ela_vs_dela}b and Figure~\ref{fig:boxplot_snr}) with high importance. Hence, feature selection becomes less important --- though still potentially valuable in some cases\added{, see \citet{seiler2024synergies}} --- as will be demonstrated in the experiments (see Section~\ref{sec:exp}). 
To achieve this, a self-supervised learning task is employed, which educates the model to formulate a representative and unique feature vector for an optimization instance -- a process often termed \textit{feature-learning}. 
Importantly, Deep-ELA is not restricted to single-objective optimization problems and can naturally be applied to multi-objective optimization problems. We will showcase Deep-ELA's competitive edge against the traditional ELA but also feature-free approaches in existing studies \citep[e.g.,][]{seiler2022collection, prager2022automated}. We also applied our model to a bi-objective case analogous to \citet{rook2022potential}. See Section~\ref{sec:exp} for an outline of all the case studies.

To the best of our knowledge, only \added{two} alternative approaches, termed DoE2Vec~\citep{van2023doe2vec} \added{and TransOpt~\citep{cenikj2023transopt}}, parallels our proposed method. However, DoE2Vec varies substantially from Deep-ELA: its model, rooted in a basic autoencoder design, solely focuses on the objective space $\spaceY$, completely disregarding the decision space $\spaceX$. This reduced informational scope results in a markedly inferior performance compared to the approaches by \citet{seiler2022collection}. Additionally, the proposed method is not invariant to the order of objective values in $\spaceY$, potentially affecting its efficacy. \added{Further, \citet{cenikj2023transopt} propose also Transformer models, TransOpt, that were trained on an AAS task but the learned (hidden) features can also be used as supplementary for classical ELA features. Contrary to Deep-ELA, the learned features are specific to the underlying AAS task and may lack generalization to other downstream tasks.}

In terms of Deep-ELA's architecture (refer to Figure \ref{fig:transformer_layout}), we predominantly adhered to the traditional transformer encoder blueprint as suggested by \citet{vaswani2017attention}. Contrary to the PCT design by \citet{guo2021pct} and \citet{seiler2022collection}, we have embedded extra feed-forward modules to align more closely with the original transformer. \added{According to a recent study \citep{geva2020transformer}, the feed-forward layers in transformers act like a memory that stores learned patterns; lower layers store more basic patterns while higher layers store more complex ones. In our case, we expect Deep-ELA to learn and memorize certain patterns of optimization problems.} Subsequently, we adopted pre-normalization as recommended by \citet{nguyen2019transformers} and used \textit{Gated Linear Units}~\citep[GLU;][]{dauphin2017language} as the activation function between all feed-forward modules to reduce the training time \added{as demonstrated by the authors}. For the \textit{feature extractor} (the last linear layer), we opted for the \texttt{Tanh} activation, mapping all activations to $[-1,1]$ and, hence, dispensing the need for feature normalization in later tasks.

The final model, \textit{Deep-ELA}, can be formally expressed as
\begin{align}
    \text{Deep-ELA} \colon (\spaceX, \spaceY) \rightarrow \spaceDELA
\end{align}
with $\nDELA$-dimensional feature space $\spaceDELA \subseteq \R^\nDELA$. Analogous to ELA, Deep-ELA requires a sample of $(X, Y) \in (\spaceX, \spaceY)$ as input. The resulting features lie within $[-1,1]$ and exhibit minimal correlation and redundancy. This contrasts with features from $\spaceELA$, which are un-normalized and often redundant (see Figure~\ref{fig:corr_ela_vs_dela} for a comparison).

During the training phase, $X$ is drawn randomly uniformly within the bounded decision space $\spaceX$. We favored uniform sampling over quasi-random sampling strategies to \textit{(1.)} reduce additional computational overhead during the training procedure, \textit{(2.)} reduce Deep-ELA's dependence on a particular sampling method, and \textit{(3.)} enhance the challenge of the self-supervised training task (which is described in more detail in Section~\ref{sec:loss}). 
After pre-training, more advanced sampling methods, such as LHS and Sobol sampling, can be employed to improve coverage of the decision space.

Deep-ELA can accommodate any sample length of $(X,Y)$ but has a constraint regarding the sum of the decision and objective space dimensions: $d + m \leq \nu$. Here, $\nu$ is an additional hyperparameter of Deep-ELA, termed the \textit{degree of dimensionality}. It specifies the maximum combined dimensionality of $\spaceX$ (denoted by $d$) and $\spaceY$ (denoted by $m$). Deep-ELA is versatile and can handle varying dimensionalities $d \in [1, \nu]$ and $m \in [1, \nu]$, 
provided their combined dimensionality does not surpass $\nu$. This flexibility extends to multi-objective optimization problems since the model can accept $m \geq 1$.  It is worth noting that the model was trained on problem instances with $d \geq 2$ and $m \geq 1$. 


\subsection{Model Structure}

\begin{figure}[tb]
    \centering
    \input{figures/transformer_layout}
    \caption{Illustration of the chosen topology of the backbone model without the training heads. The model receives $(\spaceX,\spaceY)$ as input and outputs $\spaceDELA$ -- and, optionally, the embedding of the tokens $\spaceT_\text{Final}$ after the final LayerNorm. $\spaceT_\text{Final}$ is only relevant for the contrastive loss during training and is ignored after training. 
    The initial $k$NN embedding \added{\citep{seiler2022collection}} is 
    used to capture the local information of all points from the input sample, and  
    followed by a stride operator to optionally reduce the number of tokens without losing information.
    Next, the model consists of six Multi-Head Attention blocks, followed by a Feed-Forward block of two successive Linear layers each. The LayerNorm layers are after the shortcuts as proposed by \citet{nguyen2019transformers}. We chose GLU activations in the Feed Forward layers with a $4\times$ larger number of hidden neurons.
    The last Linear + GLU layer projects the high-dimensional embeddings into lower dimensions. Afterward, the mean over all tokens is computed and normalized into $[-1,1]$ by a Tanh activation.}
    \label{fig:transformer_layout}
\end{figure}
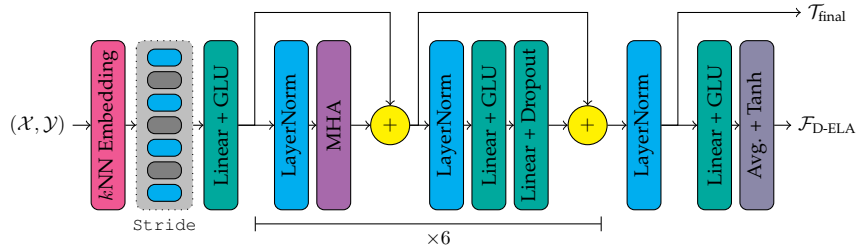

Our proposed Deep-ELA model is depicted in Figure~\ref{fig:transformer_layout}. The workflow begins with the input $(X,Y) \in (\spaceX, \spaceY)$, 
which undergoes a \textit{$k$-Nearest-Neighborhood} ($k$NN) embedding as illustrated in Figure \ref{fig:knn_embedding}. This embedding methodology, originally introduced by \citet{seiler2022collection}, seeks to incorporate the local neighborhood of every $\vec{x} \in X$. Given that transformers excel at capturing global patterns, but may overlook local nuances, the $k$NN embedding ensures local information is not ignored. Post-embedding, the encoding undergoes a projection to the deep learning network's hidden number of features, typically denoted as $d_\text{model}$ (refer to Section \ref{sec:kk_embedding}). Subsequently, every member of $\vec{x} \in X$ alongside its $k$ nearest neighbors is termed as \textit{tokens}, aligning with the terminology conventionally associated with transformers.

Following this linear projection, the sequence of tokens is sequentially processed by six combined modules of \textit{Multi-Head Attention} (MHA) and \textit{Feed-Forward} (FF), concluding with a final \textit{Layer Normalization}~\citep[LN;][]{ba2016layer}. Through experimentation, we determined that six iterations strike an optimal balance between model intricacy and performance. \added{This choice was also supported by \citet{vaswani2017attention} as they also opted for six iterations.} 
The output post-LN layer manifests as a normalized $(n \times d_\text{model})$-dimensional tensor, with $n$ representing the number of tokens. To derive the features $\spaceDELA$, the final embedding is condensed via a linear layer with GLU activation, referred to as \textit{feature-extractor}, into a $\R^{n \times \nDELA}$ tensor. Ultimately, a mean pooling step across the token sequence yields $\nDELA$ features, which are constrained within $[-1,1]$ by a concluding Tanh activation. 
An illustration of the model's topology and an overview of its associated hyperparameters are given in Figure~\ref{fig:transformer_layout} and Table~\ref{tab:model_hps}, respectively.

\begin{figure*}[tb]
    \centering
    \input{figures/training_heads}
    \caption{The two training heads on top of the backbone model. Note that the first two layers before the student's head are part of the backbone model (\textit{from BB.}). Both heads are removed after training. The student's head gets updated by gradient descent while the momentum head is an old version of the student's head and gets updated through EMA. The design follows closely the idea of \citet{chen2021mocov3}.}
    \label{fig:training_heads}
\end{figure*}
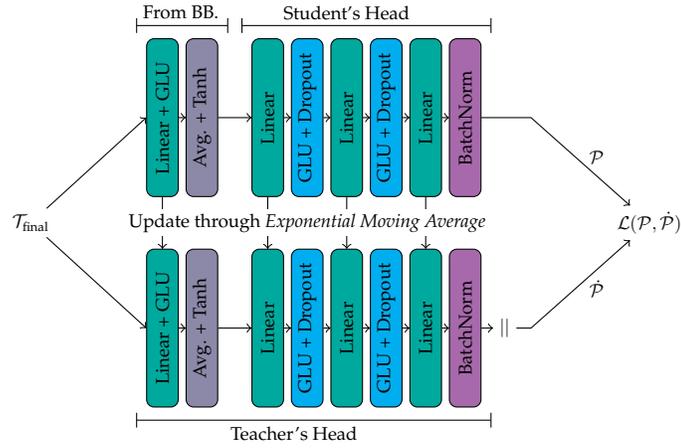

This architecture termed the \textit{backbone model}, is versatile for diverse downstream tasks. However, as the backbone essentially functions as a feature generator, it lacks trainability. Consequently, to facilitate training, we incorporate \textit{(1.)} two \textit{feature-heads} (distinguishing one as a so-called \textit{student} and the other one as a \textit{teacher}), 
and \textit{(2.)} an additional teacher feature-extractor, as visualized in Figure~\ref{fig:training_heads}. 
The student components are updated via \textit{gradient descent}, while the teacher counterparts are updated via \textit{exponential moving average}~(EMA). 
Both are integral for the self-supervised loss computation. The training loss strategy revolves around providing distinct, augmented versions of the same objective instance to both the teacher and student. Here, the teacher generates target projections from which the student gleans insights. However, we refrained from using two backbone models (one for the student and one for the teacher) as this would drastically increase the total amount of model parameters. So, the backbone model is the same for both the student and teacher. More details about the loss function and the training routine are provided in Section~\ref{sec:loss}. 

Moreover, our approach employs the \textit{InfoNCE}~\citep{oord2018representation} loss function, specifically tailored for self-supervised learning. Notably, our model is strongly influenced by the principles of \textit{Momentum Contrast for Unsupervised Visual Representation Learning}~\citep[MoCo;][]{chen2021mocov3} in its third iteration. Diverging from MoCo V3, we abstained from using a concluding MLP predictor, having observed its detrimental impact on performance and convergence in our context. Instead, our choice was a concluding \textit{Batch Normalization}~\citep[BN;][]{ioffe2015batch} with running mean and standard deviation but without the scale and shift parameters as we aim for $z$-standardization of the output projection. Even though the Batch Normalization layer did not notably elevate our performance, it amplified differences between dissimilar functions within the feature space, making distinctions more pronounced. 

Given their pivotal roles in our model, the $k$NN embedding layer and the contrastive loss function will be described in more detail in the subsequent sections.

\subsubsection{$k$NN Embedding}
\label{sec:kk_embedding}

\begin{figure*}[tb]
    \centering
    \input{figures/knn_embedding}
    \caption{For every $\vec x_i \in X$ in the decision space, the $(k-1)$ nearest neighbors are identified and together with their respective objective values combined into a single vector. Next, the $(k-1)$ nearest neighbors are projected into the local neighborhood but centered by $\vec x_i$ and $\vec y_i$, i.e., subtracting them from all its $(k-1)$ nearest neighbors. We refer to $(\vec x_i, \vec y_i)$ as the global context and to all its $k-1$ neighbors as the local context.}
    \label{fig:knn_embedding}
\end{figure*}
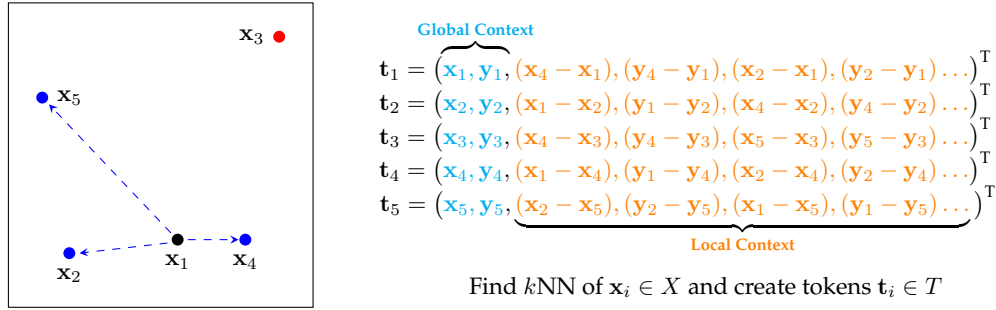

Our Deep-ELA model, schematically represented in Figure~\ref{fig:transformer_layout}, operates in two steps: input processing and embedding. Both are outlined in the following.

\paragraph{Input Processing:}

The input sets $(X,Y)$ undergo individual $z$-standardization, both per set and per dimension. To extend their dimensionality, they are padded with zeros resulting in $X', Y' \in \R^{n \times \nu}$. They are then combined into a single set, $\dot T = X' \,||\, Y'$, where $||$ indicates the element-wise concatenation operator. Following this procedure, $\dot T \subseteq \R^{n \times 2\nu}$ 
is structured such that its first $\nu$ dimensions are always decision values, while the subsequent $\nu$ dimensions represent objective values. Although this approach results in a matrix where at least half of the entries are zeros (given the dimensionality constraint $d + m \le \nu$), it offers two advantages compared to using an additional indicator vector that discriminates between decision and objective values: \textit{(1.)} all input variables remain real-valued, eliminating a mix of numeric and categorical values (that an identifier would introduce); \textit{(2.)} adding an indicator vector of length $\nu$ would also produce an input of dimensionality $n \times 2\nu$.

\paragraph{Embedding:}

For each $(\vec x, \vec y) \in (X, Y)$, its $k-1$ nearest neighbors are identified, projected to its local neighborhood, and then concatenated to $(\vec x, \vec y)$, resulting in: 
\begin{align}
    \vec t = \big(\vec x, \vec y, (\vec x_{1} - \vec x), (\vec y_{1} - \vec y), \ldots, (\vec x_{k-1} - \vec x), (\vec y_{k-1} - \vec y) \big)^\top.
\end{align}

The tokens $\vec t \in T$ form a vector with $T \subseteq \R^{n \times 2k\nu}$.  
Every element in $(X,Y)$ is termed as the \textit{global context} due to its absolute position in the decision and objective spaces. Conversely, the $(k-1)$ nearest neighbors represent the \textit{local context}, defined by their relative positions to each $(x,y)$. This distinction is pictorially elaborated in Figure \ref{fig:knn_embedding}. The embedding process can be formally outlined as:
\begin{align}
    k\text{NN-Emb.} \colon (X,Y) \rightarrow \spaceT \subseteq \R^{n \times 2k\nu}\text{.}
\end{align}

Next, $\spaceT$ undergoes a linear projection, followed by a token-wise applied GLU activation, 
mapping the points to a higher-dimensional space $\R^{n\times d_\text{model}}$ (see Figure~\ref{fig:transformer_layout}).

\subsubsection{Contrastive Loss}
\label{sec:loss}

As previously highlighted, the model employs two distinct heads during training. The \textit{student} head updates via gradient descent, while the \textit{teacher} head employs \textit{exponential moving average}~(EMA) for updates. The outputs of these heads are termed as the \textit{online-projection} $\spaceP \subseteq \R^{8\cdot\nDELA}$ (from the student), and the \textit{target-projection} $\dot \spaceP \subseteq \R^{8\cdot\nDELA}$ (from the teacher). The eight-fold dimensionality of the projections increases the heads' flexibility in aligning their representations. This methodology takes inspiration from \citet{grill2020bootstrap}, who also considered higher dimensionalities for their heads. 

The model's training closely mirrors the principles of MoCo V3, including the adoption of the InfoNCE loss function, as introduced by \citet{oord2018representation}. It has shown state-of-the-art results in other feature-learning endeavors, as evidenced by \citet{grill2020bootstrap} and \citet{chen2021mocov3}. Given a batch of online-projections, $P=\big\{\vec p_i \in \spaceP \mid\forall i \in \{1,\ldots,j\}\big\}$, and a batch of target-projections, $\dot P=\big\{\dot{\vec p}_i \in \dot\spaceP \mid\forall i \in \{1,\ldots,j\}\big\}$, with $j$ being the batch-size, the overall goal of the InfoNCE is to fulfill the following equation:
\begin{align*}
    \sigma(P\dot P^\top) \overset{!}{=} I\text{.}
\end{align*}
Here, $\sigma$ is the \texttt{Softmax} activation, and $I$ is the identity matrix. Broadly speaking, the loss function aims to maximize the covariance between an instance's online- and target-projection and to minimize it between different instances. The InfoNCE loss function is formulated based on the \textit{cross-entropy}:
\begin{align}
    \mathcal{L}_\text{InfoNCE}(\spaceP, \dot \spaceP; \tau) = 2\tau \cdot H\biggl(\sigma\bigg(\frac{\spaceP \times \dot \spaceP^\T}{\tau}\bigg)\biggr) + 
    D_\text{KL}\biggl(\sigma\bigg(\frac{\spaceP \times \dot \spaceP^\T}{\tau}\bigg), I\biggr)\text{.}
\end{align}
In this equation, $H$ and $D_\text{KL}$ denote the \textit{entropy} and the \textit{Kullback-Leibler divergence}, respectively. The cross-entropy approach provides pronounced gradients, even for vanishing activations, compared to the e.g. \textit{mean squared error}~(MSE). The loss function also depends on the critical hyperparameter $\tau$, which is the temperature of the \texttt{Softmax} activation $\sigma$. Usually, the value of $\tau$ is set in the range $(0,0.3]$. The smaller $\tau$ is, the higher the loss for \textit{hard negative samples}, i.e., distinct instances that yield similar projections and, thus, are \textit{hard} for the model to differentiate. On the contrary, for larger values of $\tau$, the loss is more evenly spread over hard and easy negative samples. Through initial testing, we found that $\tau=0.05$ works well in our scenario. \added{As an explanation, we argue that small $\tau$-values work better in our scenario as we rely solely on the stochasticity of the synthetic data generator. We cannot control the characteristics of the generated instances. So it is very likely that most instances are easy to distinguish and hard negatives occur rather infrequently. To account for this, we opted for two solutions: (1)~a small $\tau$-value will penalize hard negatives effectively also in large batches and (2)~utilizing large batches to increase the likelihood of the appearance of hard negative samples within a single batch.} 

During training, the model receives two augmented pairs of $(X_1,Y_1), (X_2,Y_2)$ for every problem instance $Z$ in a batch of $j$ problem instances. For every batch, the student and also the teacher process both augmentations of every instance, yielding four distinct projections: $P_1, P_2 \in \spaceP$ (from the student) and $\dot{P}_1, \dot{P}_2 \in \dot\spaceP$ (from the teacher). The overall loss is computed as:
\begin{align}
    \mathcal{L} = \frac{1}{2} \big( \mathcal{L}_\text{InfoNCE}(P_1, \dot P_2; \tau) + \mathcal{L}_\text{InfoNCE}(\dot P_1, P_2; \tau) \big)\text{.}
\end{align}
The deliberate pairing of $(P_1, \dot P_2)$ and $(\dot P_1, P_2)$ aims to maximize the loss across both augmented versions. For augmentation, we adopted rotations and inversions of decision variables and randomized the sequence of decision and objective variables. As the augmentations do not alter the underlying optimization problem, the predicted features should be invariant to these changes. Furthermore, the $z$-standardization in the embedding layer ensures the model remains invariant to scale and shift modifications.

\section{Datasets}\label{sec:Data}
In total, we considered four sets of optimization problem suites for training: single-objective problems from the \textit{Black-Box Optimization Benchmarking Suite}~\citep[BBOB;][]{Hansen2009bbob}, multi-objective instances from the \textit{Biobjective-BBOB Suite}~\citep[Bi-BBOB;][]{brockhoff2022bibbob}, selected instances from the R-package \texttt{smoof}\footnote{https://github.com/jakobbossek/smoof} (\textit{Single- and Multi-Objective Optimization Test Functions}), and an adapted version of the random function generator as introduced by \citet{tian2020recommender}. 

\subsection{Black-Box Optimization Problems}
\label{sec:bbob}
Most validation and testing optimization problems come from BBOB, provided by the \textit{COmparing Continuous Optimisers}~\citep[COCO;][]{nikolaus2019coco} platform. COCO offers a diverse range of continuous optimization problem suites, spanning from single- and multi-objective to noisy and mixed-integer problems. This study primarily engaged with the 24 noiseless, single-objective functions from the (classical) BBOB suite. 
For each of these functions, $1\,000$ unique instances were produced, of which the first 500 served for testing and the latter for validation purposes (see Section~\ref{sec:exp}). 
Function dimensionalities were chosen from $d \in \{2,3,5,10\}$. Although the suite includes functions with $d \in \{20,40 \}$, our model's current dimensional constraints prevent their inclusion.

For the multi-objective case study, we analyzed the same multi-objective problems that \citet{rook2022potential} examined. These bi-objective ($m=2$) functions have a decision space dimensionality of $d=2$. Specifically, we considered the Bi-BBOB functions $f_{46}, f_{47},$ and $f_{50}$, along with the ZDT, DTLZ, and MMF function groups (excluding ZDT5 and MMF13) from the \texttt{smoof} package. Further details are given in Section \ref{sec:mo:aas}.

\subsection{Random Optimization Problems}
\label{sec:random}
\begin{figure}[tb]
    \subfloat[\centering Dispersion]{{\includegraphics[height=41mm, trim={45 0 20 23px}, clip]{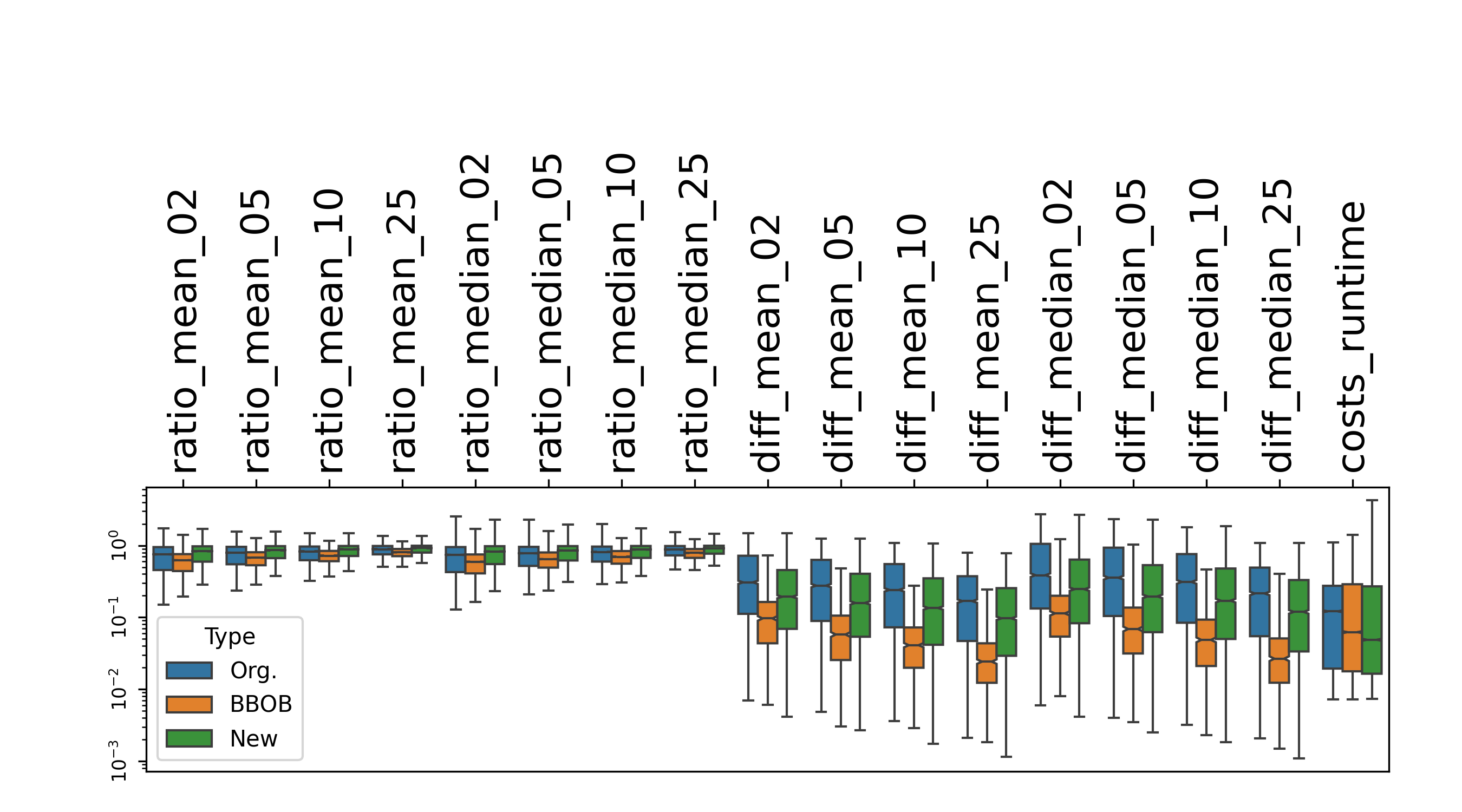}}}
    \subfloat[\centering Meta]{{\includegraphics[height=41.5mm, trim={0 0 0 23px}, clip]{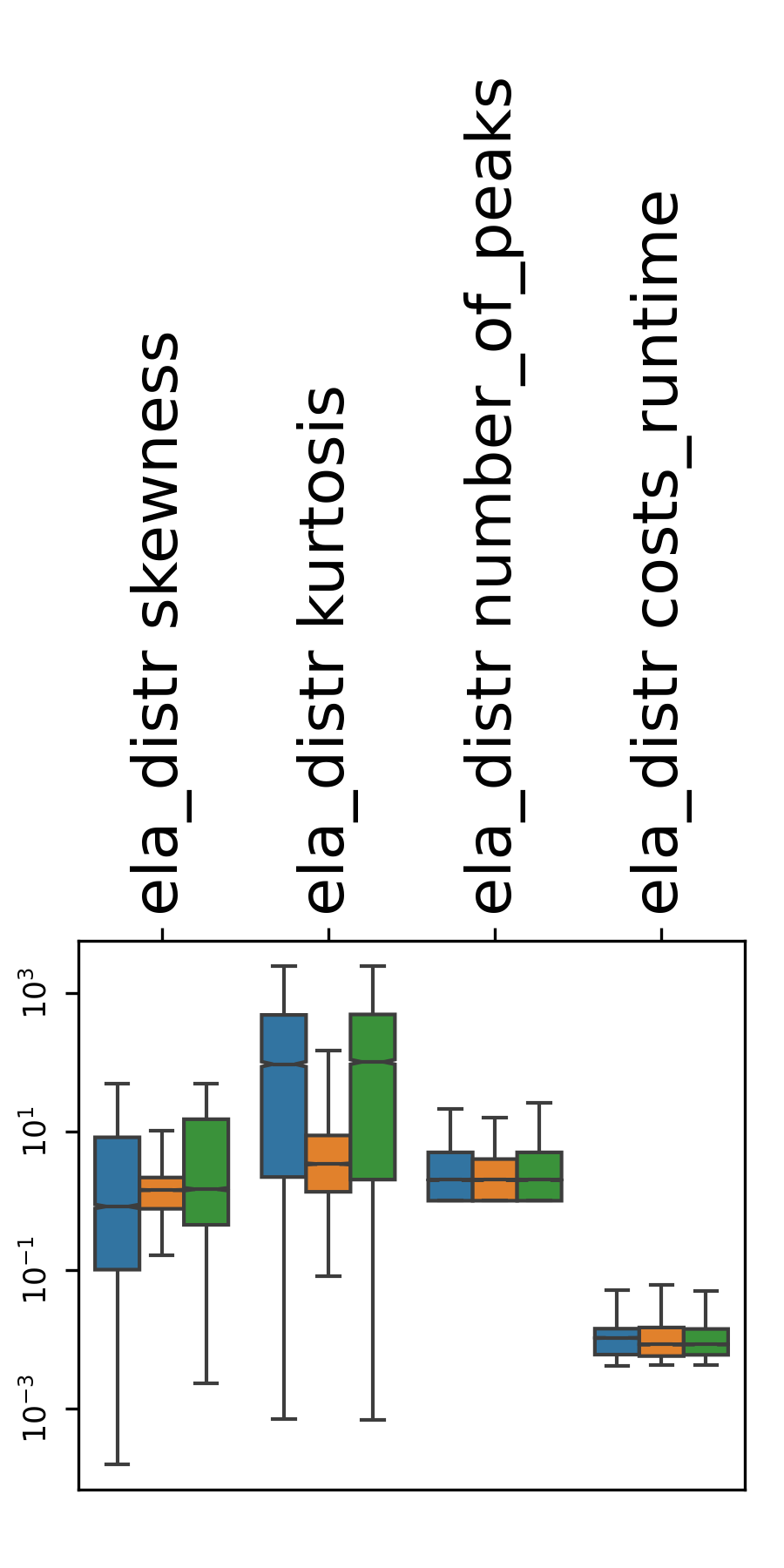}}}
    \subfloat[\centering $Y$-Distribution]{{\includegraphics[height=41mm, trim={15 0 10 23px}, clip]{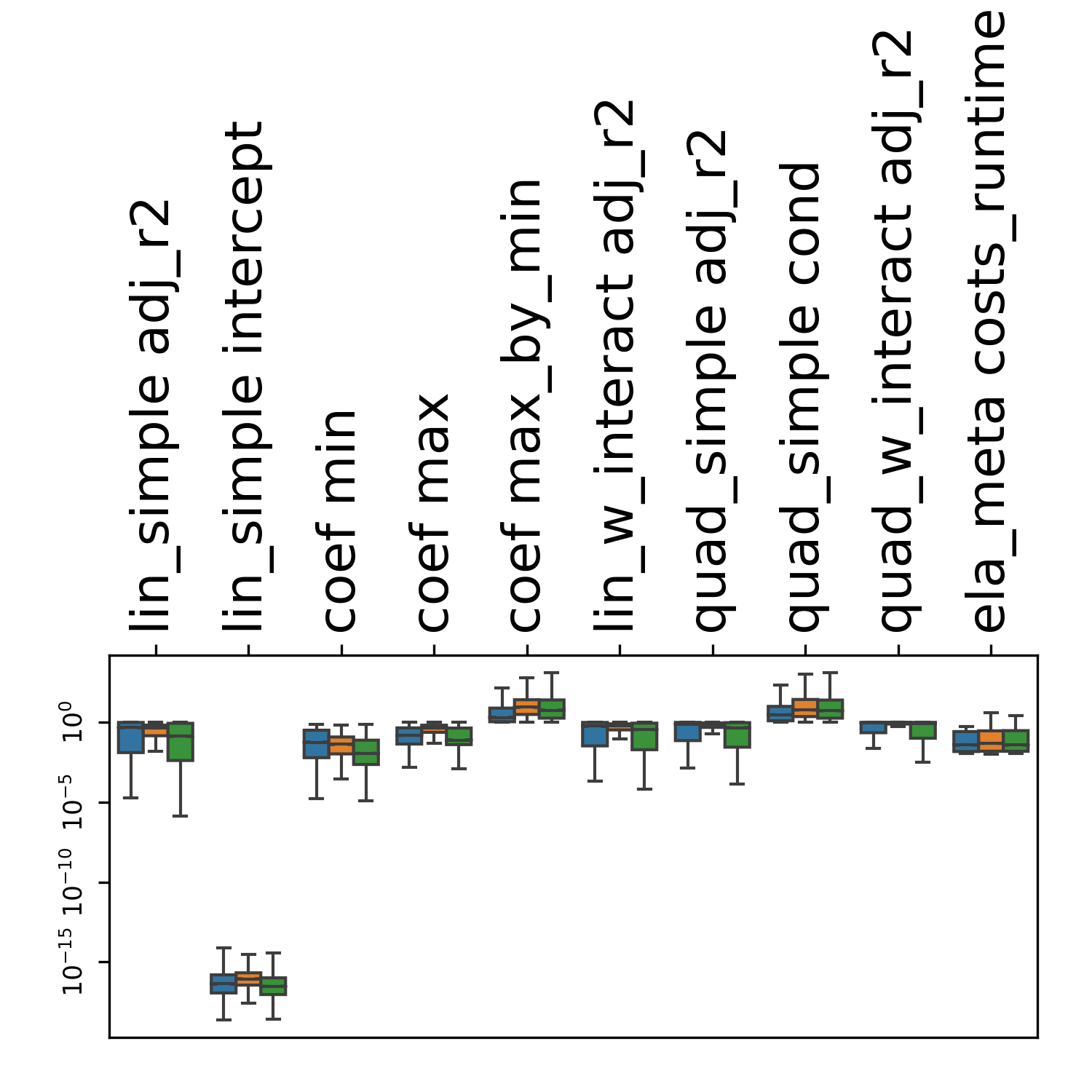}}}

    \caption{Distributions of feature values for three exemplary sets of ELA features, differentiated by three types of test problems -- the original instance generator by \citet{tian2020recommender} (blue), the BBOB suite (orange), and our adapted generator (green). For all problems, we used dimensions $2,3,5,10$. In the case of BBOB, we considered all 24 noiseless functions with instances $1$ to $20$. For the random generators, we used the same number of instances (in total).} 
    \label{fig:bbob_vs_random}
\end{figure}
\citet{tian2020recommender} introduced a method capable of generating an arbitrary number of optimization problems in milliseconds. This approach hinges on a random tree-like structure constituted of various operators such as \texttt{mean}, \texttt{sum}, \texttt{exp}, \texttt{log}, among many others. This generator operates on three input parameters: the number of dimensions of the decision space ($d$), and the lower and upper bounds of the number of operators. 
Once initiated, the generator fabricates an optimization problem for a $d$-dimensional set of decision variables $\spaceX$ by randomly assorting a number of operators bounded by the provided lower and upper bounds. The result is a randomly constructed optimization problem that yields the objective values $Y$. We manually calibrated \added{the upper and lower bounds of} the generator\deleted{'s hyperparameters} to craft optimization 
problems that are similar to BBOB. \added{We found that when choosing these bounds too small or too large, the generator produces objectives that are too simple or too complicated in comparison to BBOB. Yet, as the models (like any other machine learner) can only generalize to instances that are similar to the training data, we need to ensure that BBOB is not out-of-distribution to the models. We found that upper and lower bounds of $(4,32)$ fulfill this criterion well.} Additionally, we incorporated extra operators to amplify the linear and squared dependencies among decision and objective variables, noting that the original generator somewhat diverged from the BBOB suite in these aspects. Furthermore, we tweaked the skewness and kurtosis of the objective variables for closer alignment with BBOB (refer to Figure~\ref{fig:bbob_vs_random}). \added{Last, the original \texttt{Python} implementation yielded \texttt{NaN}-values quite often. We identified and modified those operators that cause the appearance of \texttt{NaN}-values. Afterward, we observed that the generator produces diverse optimization problems that are more similar to BBOB in terms of their complexity of ELA features (see Figure~\ref{fig:bbob_vs_random}). Yet, it shall be noted that the random functions are not identical to BBOB. Instead, the random functions cover similar characteristics as BBOB.}

However, given the inherent randomness of the generator, not all produced optimization problems are applicable for training. We thus dismiss any instance that does not satisfy the following three conditions:
\begin{enumerate}
    \itemsep0em 
    \item The output $Y$ must be a set of scalars.
    \item The objective values must have a standard deviation of at least $0.1$
    \item The objective values must lie within $[-10\,000\,000, \; +10\,000\,000]$.
\end{enumerate}

To generate multi-objective instances, we randomly designed $m$ single-objective functions and combined them into a multi-objective instance. While this methodology proved satisfactory for our purposes, the combined set of random single-objectives may not fully encapsulate the inherent traits of multi-objective problems. Notably, such problems often harbor interacting objectives, potentially exhibiting mutual dependencies. However, randomly creating multi-objective instances with such interdependencies proved to be much more challenging. We thus postponed this endeavor for future work, and hope that randomness will generate objectives with mutual dependencies. We built upon the Python implementation\footnote{https://www.github.com/Basvanstein/doe2vec/tree/main/src/modulesRandFunc} by \citet{van2023doe2vec}.

\section{Experiments}\label{sec:Exp}
\label{sec:exp}
In the following sections, we describe and analyze the pre-training of our Deep-ELA models. Subsequently, we examine the results of three case studies. The first two, \textit{High-Level Property Prediction}~\citep{seiler2022collection} and \textit{Single-Objective Automated Algorithm Selection}~\citep{prager2022automated}, have been conducted in previous studies. Please note that we directly took the results from previous work of \citet{seiler2022collection,prager2022automated,van2023doe2vec}. 
The third case study, i.e., the \textit{Multi-Objective Automated Algorithm Selection}, is adapted from \citet{rook2022potential} and introduces a newly created dataset.

\subsection{Pretraining}
\label{sec:pretraining}

\begin{table}[tb]
    \centering
    \caption{Hyperparameter settings of our models. In total, we trained four final models, which were configured systematically based on a pre-study. The number of randomly generated training instances is given per epoch, so, every model was trained on a total of 250\,000\,000 randomly generated single- and multi-objective functions.}
     \input{tables/model_hps}
    \label{tab:model_hps}
\end{table}

Before training our final models, we conducted several preliminary studies to determine the optimal hyperparameters and model topologies for our use cases. We experimented with various $\tau$ values, momentum factors, numbers of heads, and layers. Eventually, we systematically selected the hyperparameters presented in Table~\ref{tab:model_hps}. 

Given the availability of two high-performance computing (HPC) servers -- one with three NVidia Quadro RTX 6\,000 GPUs and the other with three NVidia RTX A 6\,000 GPUs -- we decided to train the larger models on the RTX A 6\,000 GPUs, which have 48GB of GPU memory, compared to the Quadro's 24GB. The smaller models were comfortably trained on the 24GB GPUs. We maximized the batch size, retaining a small buffer, to include as many samples as possible within each batch as the used loss function benefits from large batches. Each model was trained across three GPUs, handling three batches concurrently. After the forward pass, the three batches were merged into one joint batch for loss computation, effectively tripling the batch size and increasing the likelihood of the appearance of hard negative samples. 

We chose to train both a medium- and a large-sized model. The medium model can manage a total dimensionality of six, while the large model can accommodate up to twelve total dimensions. The large model has twice the width of the medium model and produces twice the number of features (48 vs. 24). Due to the squared growth of transformer complexity with the number of tokens, we introduced a stride factor. A stride of two implies that every second token is omitted \textit{after} the $k$NN-embedding, as depicted in Figure~\ref{fig:transformer_layout}. Consequently, the large model's complexity is reduced to a fourth, compared to a model with a stride of one, maintaining a larger batch size during the training of the large model. We then trained each model-variant with sample sizes of $25d$ and $50d$, resulting in four final models\added{, where $d$ stands for the dimensions of $\spaceX$. Hence, the sample-size scales linearly with the number of dimensions of the decision space.} For the $50d$ models, we reduced the batch size by half but compensated with gradient accumulation over two iterations. However, InfoNCE does not benefit much from gradient accumulation since its efficiency relies on a substantial number of samples. Still, we opted for gradient accumulation to maintain a consistent number of update steps and training instances, as seen with the $25d$ models.

For training, we employed \texttt{PyTorch-Lightning}\footnote{https://lightning.ai/} in tandem with \texttt{PyTorch}\footnote{https://pytorch.org/}. Post-training, we removed the two heads necessary during the training phase, resulting in a versatile backbone model suitable for various downstream tasks. Subsequently, we incorporated classical machine learners atop this backbone. No further fine-tuning of the backbone model was undertaken or deemed necessary. \added{Instead, the backbone model can be directly applied as the considered studies have similar boxed constraints for $\spaceX$ as the training instances. Yet, if e.g. the box constraints are noticeably dissimilar to the training set ($[-5,5]$) fine-tuning may be advisable. But this requires further testing in future work.} \deleted{However, it is possible to fine-tune the backbone model for different tasks, although this demands a deep understanding of deep learning techniques.} To emulate a user proficient in general machine learning but not deeply versed in deep learning specifics, we relied exclusively on \texttt{Scikit-Learn} and refrained from fine-tuning the backbone model. Further, we did not apply feature selection to demonstrate that the instance features generated by the Deep-ELA approach contain features of high relevance and little redundancy. \added{Last, it should be noted that the backbone model did not receive any instances from the BBOB suite or similar frameworks and was solely trained on randomly generated instances. Further, the model receives each optimization instance exactly once. Hence, we argue that any overfitting to existing optimization problems is hardly possible. More importantly, as the models were not trained to predict certain labels as it is usually done in supervised learning, they cannot just remember labels for a specific task.}

\subsection{High-Level Property Prediction}
\label{sec:hl2p}
\begin{table}[tb]
    \centering
    \caption{Assignment of the 24 single-objective noiseless BBOB functions into three frequently used \textit{High-Level Properties}~(HLP), as outlined and proposed in \citet{mersmann2010benchmarking} and \citet{kerschke2015detecting}.}
    \label{tab:hlp}
    \input{tables/bbob_hlp}
\end{table}

The term \textit{High-Level Properties}~(HLP) refers to a collection of structural attributes describing the fitness landscape of the examined single-objective optimization. These properties enable the characterization of the problem -- e.g., into low, medium, and high multimodality -- facilitating the identification of (dis-)similar problems. This holds especially true for high-dimensional problems, which cannot be visually represented or analyzed. Moreover, 
these properties are pivotal for algorithm selection, algorithm configuration, and for crafting algorithms tailored to specific objectives. 
%
%
Of the eight properties detailed in \citet{mersmann2011exploratory} and \citet{kerschke2015detecting}, the following three exert the most pronounced influence on optimization problem complexity:
\begin{compactenum}
    \itemsep0em 
    \item \textbf{Multimodality}: The \textit{degree of multimodality} aggregates the number of local optima into `low', `medium', and `high' multimodality, as well as `none' in case of unimodal problems.
    \item \textbf{Global Structure}: Chronicles the \textit{structural} link between local and global optima.
    \item \textbf{Funnel}: Signifies the presence of a \textit{funnel}-like layout of the local and global optima.
\end{compactenum}
The assignment of the 24 BBOB functions into the three HLPs is given in Table~\ref{tab:hlp} for convenience. For predicting the HLPs, we adhered to the setup prescribed in \citet{seiler2022collection}. This entailed evaluating the 24 BBOB functions, adopting instances $\{1, \ldots, 100\}$ for training and $\{125, \ldots, 150\}$ for testing, with $d \in \{2,3, 5, 10\}$. We juxtaposed our findings against those from \citet{seiler2022collection}. Additionally, we consider the results reported by \citet{van2023doe2vec} in the HLP study of their Doe2Vec method. For prediction purposes, we utilized \textit{Random Forests}~(RF) and \textit{Support Vector Machines}~(SVM) to forecast the three HLPs based on features generated by the pre-trained backbone model. 
The obtained (and collected) results are listed in Table~\ref{tab:results_hlp}.

\begin{table}[tb]
    \centering
    \caption{F1-Score with macro aggregation (F1-Scores a computed individually per class and then \texttt{mean}-aggregated over all classes) of predicting \textit{High-Level Properties} of all 24 BBOB functions and instances $126$ to $150$. The results marked with a $^{*}$ are directly taken from \citet{seiler2022collection} while the ones marked with $^{**}$ are taken from \citet{van2023doe2vec}. The last eight columns list the performances obtained by our proposed Deep-ELA models. \citet{van2023doe2vec} did not provide results for $d=3$ and aggregated over all dimensions (all). Our medium models cannot handle data with more than six total dimensions. Therefore, there are no results for $d=10$ for the medium models.} 
    \label{tab:results_hlp}
    \input{tables/results_hlp}
\end{table}
As summarized in Table~\ref{tab:results_hlp}, the performances of our models lag behind those of \citet{seiler2022collection} but are superior to the ones listed in \citet{van2023doe2vec}. 
This was somewhat anticipated, given that \citet{seiler2022collection} explicitly trained on BBOB instances for HLP prediction. In contrast, both \citet{van2023doe2vec} and our approach were trained by self-supervised learning on autonomously generated instances. Notably, our medium-scale models often outperformed their larger counterparts. Two potential reasons underpin this trend. Firstly, the large models, having been utilized with a stride of two, bypass half of the accessible data, potentially decreasing their efficacy. The second rationale suggests that while the larger models underwent an identical number of training iterations, they were exposed to a broader range of dimensions, possibly hindering their ability to generalize equally well to specific dimensions. Lastly, our models with $50d$ consistently 
outperformed 
the $25d$ models, echoing the findings of \citet{seiler2022collection}, who observed superior performance from transformers with a sample size of 500 versus those with a size of 100. In terms of the considered machine learning models, our experiments did not discern any substantial performance discrepancy between the RF and SVM results; both exhibited roughly comparable efficiency.

\subsection{Single-Objective Automated Algorithm Selection}
\label{sec:so:aas}
In our subsequent case study, we both replicated and expanded upon the experiments presented by \citet{prager2022automated}. Integral to this study was the data originally compiled and introduced by \citet{kerschke2015detecting}, encompassing benchmark results of twelve competitive, complementary algorithms -- a comprehensive list of these algorithms can be found in \citet{kerschke2015detecting} and \citet{prager2022automated}. The benchmark performances were obtained from COCO, a repository archiving the outcomes of optimizer runs from a myriad of competitions hosted on the BBOB suites. In the methodology laid out by \citet{kerschke2015detecting}, a threshold of $0.01$ was deemed acceptable, i.e., an algorithm that procures a solution within $0.01$ of the actual optimum (in objective space) is categorized as successful. An algorithm's efficacy is gauged through the \textit{relative Expected Running Time} (\textit{relERT}) metric 
\citep{auger2005performance}.
The relERT provides a relative variant of the ERT itself, by juxtaposing the ERT of a given algorithm against that of the \textit{Virtual Best Solver} (VBS). Hence, a relERT of $r$ indicates that the examined algorithm needs $r$ times as many function evaluations compared to the best algorithm from the considered portfolio. 
Note that when predicting the top-performing algorithm, the size of the initial sample must be factored in as a cost, which is first added to the algorithm's ERT prior to deriving the relERT. The sample is essentially needed for the model to scrutinize a given problem instance (in feature-free algorithm selection) and predominantly for feature computation (in feature-based algorithm selection). 

\begin{table}[tb]
    \centering
    \caption{Comparison of the relative ERT (relERT) values for the algorithm selection study on instances $1$ to $5$ of the 24 single-objective BBOB functions. The first five columns (marked with $^{*}$) are directly taken from \citet{prager2022automated}, while the last eight show the performance of our proposed Deep-ELA approach. Our medium models cannot handle data with more than six dimensions. Therefore, there are no results for $d=10$ for the medium models.}
    \label{tab:results_so_as}
    \input{tables/results_so_as}
\end{table}

Our experimental setup closely mirrored that of \citet{prager2022automated}, employing the 24 noiseless BBOB functions, instances $\{1, 2, 3, 4, 5\}$ with ten repetitions each, and $d \in \{2, 3, 5, 10\}$. In line with \citet{prager2022automated}, we executed five-fold cross-validation across the five instances for each function. 
Further, \citet{prager2022automated} employed a cost-sensitive loss function, but it was exclusively compatible with machine learners amenable to gradient descent training. This discrepancy is evident when comparing the RF and MLP models, both trained on ELA features. While the RF model significantly underperformed in relation to the \textit{Single Best Solver} (SBS), the MLP exhibited state-of-the-art results. Our research, however, sidestepped the loss function delineated by \citet{prager2022automated}, as we restricted ourselves to the \texttt{sklearn} framework, avoiding other platforms. In contrast, \citet{prager2022automated} utilized \texttt{PyTorch} complemented by a bespoke loss function and training protocol. \texttt{sklearn}, however, does not provide the functionality to implement a custom loss function or to make use of an instance-based, cost-sensitive training routine.

Diverging from the insights of our initial case study, this research unveiled a noticeable enhancement in our proposed Deep-ELA methodology compared to the results of \citet{prager2022automated}. All our models consistently outperformed the feature-free strategies delineated in \citet{prager2022automated}. In certain scenarios, our Large-$25d$ and Medium-$50d$ models even surpassed the MLP-ELA model. A consistent observation from \citet{prager2022automated} was the superior performance of the Medium-$25d$ over the Medium-$50d$ model. We postulate that the higher costs associated with acquiring supplementary information outweigh the benefits derived from this extra data. However, this paradigm does not apply to the large models, with the Large-$50d$ model being superior to the Large-$25d$ model. This trend suggests that larger models may be necessary to extract additional value from the added data of the larger sample size. Nevertheless, the disparity between the results is nuanced. Our best-performing Deep-ELA model was the Medium-$25d$ model, lending credence to the hypothesis that employing a stride of two may hamper performance. Notably, the $k$NN classifier substantially outperformed the RF classifier. In this context, we opted for a $k$ value of 69, the sole hyperparameter we carefully fine-tuned manually. 

Our investigations into single-objective AAS unveiled a pivotal revelation: Deep-ELA operates as envisaged, particularly excelling with limited datasets compared to feature-free AAS, where the propensity for overfitting makes training extensive deep-learning models a challenge.

\subsection{Multi-Objective Automated Algorithm Selection}
\label{sec:mo:aas}
Our experimental design is akin to that of \citet{rook2022potential}, albeit without the \textit{algorithm configuration} segment. Instead, we opted for the seven algorithms 
\citep[as detailed in][]{rook2022potential} in their default settings and employed an AAS approach to select the optimal algorithm. Three of these seven algorithms -- NSGA-II~\citep{deb2002fast}, SMS-EMOA~\citep{beume2007sms}, and MOEA/D~\citep{zhang2007moea} -- represent classical \textit{Evolutionary Multiobjective Optimization Algorithms}~(EMOAs), while the other four are Omni-Optimizer~\citep{deb2005omni}, MOLE~\citep{schaepermeier2022mole}, MOGSA~\citep{grimme2019multimodality}, and HIGA-MO~\citep{wang2017hypervolume}. Performance was gauged using the \textit{Hypervolume}~(HV) of the approximated Pareto front found within a budget of $20\,000$ function evaluations. For most test instances, the reference point for the HV was pre-specified. For problems without predetermined reference points (for HV computation), we derived the necessary points from the least favorable solution of all algorithms for that specific instance. To maintain HV value consistency across distinct instances, we normalized the HV values, basing them on the highest HV identified from additional runs of all algorithms with a more substantial budget of $100\,000$ function evaluations. Subsequently, the resulting normalized HV values were contrasted against the SBS-VBS gap -- and we coined this metric relative HV (relHV). In this context, a value close to zero resembles SBS performance, while a value of one relates to VBS performance. Negative values, in turn, denote performances that are inferior to the SBS. 
Formally, the metric is defined as
\begin{align}
    \text{relHV}(I,A) &= \frac{\text{HV}(I,A) - \text{HV}_\text{SBS} + 10^{-8}}{\text{HV}_\text{VBS} - \text{HV}_\text{SBS} + 10^{-8}}
\end{align}
where $10^{-8}$ is a tiny value that prevents division by zero. This could happen if the SBS and VBS are identical. 
By also including the term $10^{-8}$ in the numerator ensures that the relHV is one, in case the selector predicts the VBS.

\begin{table}[tb]
    \centering
    \caption{The mean relative HV (relHV) of our proposed Deep-ELA approach on the multi-objective AAS study. relHV values are calculated per test instance and mean-aggregated first over instances of the same function and then over functions of the same group. A value of one (zero) indicates performances comparable to the VBS (SBS). Negative values indicate a performance worse than the SBS.}
    \label{tab:results_mo_as}
    \input{tables/results_mo_as}
\end{table}

The dataset comprised ZDT, DTLZ, and MMF test instances, excluding ZDT5 and MMF13. Additionally, instances $f_{46}$, $f_{47}$, and $f_{50}$ from the Bi-BBOB were integrated. This resulted in a total of 33 instances -- 
with 
a two-dimensional decision ($d=2$) and objective ($m=2$) space each. We executed 20 repetitions per instance, of which the first 15 were used for training and the remaining five ($16-20$) for testing.

The results of our exploration are listed in Table~\ref{tab:results_mo_as}. Due to the lack of other AAS studies on multi-objective optimization, we cannot compare Deep-ELA to any other studies from the literature. A potential reason for this may be the somewhat limited applicability of ELA on multi-objective optimization problems. It is pertinent to note that the crux of this study revolves around maximizing the relHV. Given the absence of any costs to factor in for this case study -- as we measure the performance based on the found hypervolume instead of the number of function evaluations -- achieving performances on the VBS level is viable. Our findings reveal that the medium models eclipse the larger ones in performance, and the $50d$ models have a performance edge over the $25d$ models. This observation aligns with our expectations, considering that the $50d$ models harness more information than their $25d$ counterparts. We posit that the larger models are potentially hampered by the stride of two. Specifically, the large $25d$ model registered a negative relVH value for the ZDT instance group, suggesting a performance that lags behind the SBS. The $k$NN classifiers showcased a discernibly superior performance in comparison to the RF classifiers. In this context, we opted for a smaller $k=15$. Both, the Medium $25d$ and Medium $50d$ models rendered comparable outcomes when leveraging a $k$NN classifier. However, when an RF classifier was in play, the Medium $50d$ model emerged as the top performer. In conclusion, the performances achieved by our Deep-ELA approach in this study are often very good -- and, in many cases, even perfect.

\section{Discussion \& Conclusion} \label{sec:Disc}
A salient outcome of this research is the demonstrated efficacy of the Deep-ELA methodology. It can either be used out-of-the-box for analyzing single- and multi-objective continuous optimization problems, as demonstrated, or fine-tuned to various tasks on algorithm behavior and problem understanding. In comparison to the strategies delineated by \citet{prager2022automated}, Deep-ELA exhibited significant performance enhancements, particularly in its adept handling of limited datasets. Feature-free deep-learning models, like those utilized by \citet{prager2022automated}, often contend with overfitting challenges when trained on limited datasets. In contrast, Deep-ELA's design, having been trained on millions of randomly generated optimization instances, offers a remedy to such constraints. As a result, the backbone model could be integrated seamlessly into existing and novel downstream tasks without necessitating additional training or fine-tuning on the dataset from \citet{prager2022automated}. Furthermore, the derived instance features could be employed without the need for feature selection or normalization, as these features are less correlated than the commonly used ELA features, and, in addition, are all located within $[-1,1]$. 

Yet, alongside these advantages, the investigations revealed certain intricacies warranting further examination. Notably, medium-sized models demonstrated superior performance compared to their larger counterparts. 
Concurrently, a trend emerged indicating that models with a sample size of $50d$ performed better than those with $25d$. 
While the latter trend is intuitive -- a sample size of $50d$ inherently offers more information for the model to process -- the former raises questions regarding the current iteration of the Deep-ELA design. The use of a stride of two, particularly in larger models, was pinpointed as a possible bottleneck inhibiting peak performance. This observation offers a compelling avenue for further research. Even though the chosen stride of size two was indispensable for accommodating large training batches, future studies could explore more refined pooling techniques to diminish sequence sizes without significant data loss. Another potential explanation for the diminished performance of larger models may lie in their training on a more varied set of decision space dimensions. As a future direction, researchers may delve into whether extended training durations would amplify results or if other underlying factors are at play.

Despite the contained scope of the results, the Deep-ELA methodology made a noteworthy foray into the multi-objective AAS domain. The capability of Deep-ELA to seamlessly transition to multi-objective problem instances stands out as a significant achievement. This is particularly notable considering that traditional ELA features are innately tailored for single-objective problems. Moving forward, there is ample opportunity to devise more intricate studies where the Deep-ELA approach can be further tested and refined. 
At last, configuring the hyperparameters of the Deep-ELA framework in an automated manner, e.g., using efficient algorithm configurators such as irace \citep{lopez2016irace} or SMAC \citep{lindauer2022smac3}, also provides a very promising direction for future research.

\small

\bibliographystyle{apalike}
\bibliography{bib}

\end{document}

%% file: figures/aas_classic.tex
\begin{tikzpicture}[scale=0.85, transform shape]
    \node (instance) at (2,1.5) [draw, text width=2cm, fill=gray!30, align=center] {\large Problem $I \in \spaceI$};
    \node (features) at (2,0) [draw, text width=3cm, fill=gray!00, align=center] {\large Comp. Features\\ELA$\colon \spaceI \rightarrow \spaceELA$};
    \node (selector) at (6,0) [draw, text width=3cm, fill=gray!00, align=center] {\large Select\\$\dot S \colon \spaceELA \rightarrow \spaceA$};
    \node (portfolio) at (6,1.5) [draw, text width=3cm, fill=gray!30, align=center] {\large Algorithm\\Portfolio $\spaceA$};
    \node (apply) at (6,-1.5) [draw, text width=2cm, fill=gray!00, align=center] {Solve $I$\\with $A$};
    \node at (5,-2.5) [text width=6cm] {(a) Feature-Based AAS}; 

    \draw[->, line width=2pt] (instance) -- (features);
    \draw[->, line width=2pt] (features) -- (selector);
    \draw[->, line width=2pt] (portfolio) -- (selector);
    \draw[->, line width=2pt] (selector) -- (apply);
\end{tikzpicture}

%% file: figures/aas_novel.tex
\begin{tikzpicture}[scale=0.85, transform shape]
    \node (instance) at (2,1.5) [draw, text width=2cm, fill=gray!30, align=center] {\large Problem $I \in \spaceI$};
    \node (selector) at (6,0) [draw, text width=3cm, fill=gray!00, align=center] {\large Select\\$S \colon \spaceI \rightarrow \spaceA$};
    \node (portfolio) at (6,1.5) [draw, text width=3cm, fill=gray!30, align=center] {\large Algorithm\\Portfolio $\spaceA$};
    \node (apply) at (6,-1.5) [draw, text width=2cm, fill=gray!00, align=center] {Solve $I$\\with $A$};
    \node at (5,-2.5) {(b) Feature-Free AAS}; 

    \draw[->, line width=2pt] (instance.south) |- (selector.west);
    \draw[->, line width=2pt] (portfolio) -- (selector);
    \draw[->, line width=2pt] (selector) -- (apply);
\end{tikzpicture}

%% file: figures/transformer_layout.tex
\usetikzlibrary{shapes.misc, positioning}
\begin{tikzpicture}[scale=0.75, transform shape]
\node[rotate=00] at (-1.25, 0) (inp) {$(\spaceX,\spaceY)$};
\node[draw, fill=RubineRed!80, rounded corners=1mm, minimum width=3cm, minimum height=6mm, rotate=90] at (0.0,0) (knn) {$k$NN Embedding};
\node[draw, fill=Emerald, rounded corners=1mm, minimum width=3cm, minimum height=6mm, rotate=90] at (2.0,0) (ln2) {Linear + GLU};

\node[draw, fill=lightgray, dotted, rounded corners=1mm, minimum width=30mm, minimum height=10mm, rotate=90] at (1.00,0) (tokens) {};

\node[draw, fill=cyan, rounded corners=1mm, minimum width=6mm, minimum height=3mm] at (1.0,1.2) (msk1) {};
\node[draw, fill=gray, rounded corners=1mm, minimum width=6mm, minimum height=3mm] at (1.0,0.8) (tok1) {};
\node[draw, fill=cyan, rounded corners=1mm, minimum width=6mm, minimum height=3mm] at (1.0,0.4) (tok2) {};
\node[draw, fill=gray, rounded corners=1mm, minimum width=6mm, minimum height=3mm] at (1.0,0.0) (msk2) {};
\node[draw, fill=cyan, rounded corners=1mm, minimum width=6mm, minimum height=3mm] at (1.0,-0.4) (msk3) {};
\node[draw, fill=gray, rounded corners=1mm, minimum width=6mm, minimum height=3mm] at (1.0,-0.8) (msk4) {};
\node[draw, fill=cyan, rounded corners=1mm, minimum width=6mm, minimum height=3mm] at (1.0,-1.2) (tok3) {};
\node[] at (1.0, -1.75) () {\texttt{\small Stride}};

\node[draw, fill=cyan, rounded corners=1mm, minimum width=3cm, minimum height=6mm, rotate=90] at (3.25,0) (ln3) {LayerNorm};
\node[draw, fill=Purple!80, rounded corners=1mm, minimum width=3cm, minimum height=6mm, rotate=90] at (4.0,0) (ln4) {MHA};

\node[draw, fill=yellow, circle, minimum width=0.25cm] at (5.00,0) (add1) {$+$};

\node[draw, fill=cyan, rounded corners=1mm, minimum width=3cm, minimum height=6mm, rotate=90] at (6.0,0) (ln5) {LayerNorm};
\node[draw, fill=Emerald, rounded corners=1mm, minimum width=3cm, minimum height=6mm, rotate=90] at (6.75,0) (ln6) {Linear + GLU};
\node[draw, fill=Emerald, rounded corners=1mm, minimum width=3cm, minimum height=6mm, rotate=90] at (7.50,0) (ln7) {Linear + Dropout};

\node[draw, fill=yellow, circle, minimum width=0.25cm] at (8.50,0) (add2) {$+$};

\node[draw, fill=cyan, rounded corners=1mm, minimum width=3cm, minimum height=6mm, rotate=90] at (9.50,0) (ln8) {LayerNorm};
\node[draw, fill=Emerald, rounded corners=1mm, minimum width=3cm, minimum height=6mm, rotate=90] at (10.75,0) (ln9) {Linear + GLU};
\node[draw, fill=CadetBlue!80, rounded corners=1mm, minimum width=3cm, minimum height=6mm, rotate=90] at (11.50,0) (ln10) {Avg. + Tanh};
\node[] at (12.75, 0) (out_f) {$\spaceDELA$};
\node[] at (12.75, 2.) (out_t) {$\spaceT_\text{final}$};

\draw[->] (inp) -- (knn);
\draw[->] (knn) -- (tokens);
\draw[->] (tokens) -- (ln2);
\draw[->] (ln2) -- (ln3);
\draw[->] (ln3) -- (ln4);
\draw[->] (ln4) -- (add1);

\draw[->] (add1) -- (ln5);
\draw[->] (ln5) -- (ln6);
\draw[->] (ln6) -- (ln7);
\draw[->] (ln7) -- (add2);

\draw[->] (add2) -- (ln8);
\draw[->] (ln8) -- (ln9);
\draw[->] (ln9) -- (ln10);
\draw[->] (ln10) -- (out_f);

\draw[->] ([xshift=0cm]ln2.south) -- (2.6,0) -| (2.6, 2.00) -| (add1);
\draw[->] ([xshift=0cm]add1.east) -- (5.5,0) -| (5.5, 2.00) -| (add2);
\draw[->] ([xshift=0cm]ln8.south) -- (10.1,0) -| (10.1, 2.00) -- (out_t.west);

\draw[|-|](2.6, -1.75) -- (8.75,-1.75);
\node[anchor=west] at (5.5,-2) (label) {$\times6$};

\end{tikzpicture}

%% file: figures/training_heads.tex
\usetikzlibrary{shapes.misc, positioning}
\begin{tikzpicture}[scale=0.7, transform shape]
\node[] at (-1, 0) (input) {$\spaceT_\text{final}$};
\node[] at (10.75, 0) (loss) {$\mathcal{L}(\mathcal{P}, \mathcal{\dot P})$};

\node[] at (9.75, 1.25) (ff) {$\mathcal{P}$};
\node[] at (9.75, -1.25) (hf) {$\mathcal{\dot P}$};


\node[] at (1.85, 4) () {From BB.};
\node[] at (5.00, 4) () {Student's Head};
\node[] at (4.00, -4) () {Teacher's Head};

\node[draw, fill=Emerald, rounded corners=1mm, minimum width=3cm, minimum height=6mm, rotate=90] at (1.50,2) (ln00) {Linear + GLU};
\node[draw, fill=CadetBlue!80, rounded corners=1mm, minimum width=3cm, minimum height=6mm, rotate=90] at (2.25,2) (ln01) {Avg. + Tanh};

\node[draw, fill=Emerald, rounded corners=1mm, minimum width=3cm, minimum height=6mm, rotate=90] at (3.50,2) (ln02) {Linear};
\node[draw, fill=cyan, rounded corners=1mm, minimum width=3cm, minimum height=6mm, rotate=90] at (4.25,2) (ln03) {GLU + Dropout};
\node[draw, fill=Emerald, rounded corners=1mm, minimum width=3cm, minimum height=6mm, rotate=90] at (5.00,2) (ln04) {Linear};
\node[draw, fill=cyan, rounded corners=1mm, minimum width=3cm, minimum height=6mm, rotate=90] at (5.75,2) (ln05) {GLU + Dropout};
\node[draw, fill=Emerald, rounded corners=1mm, minimum width=3cm, minimum height=6mm, rotate=90] at (6.50,2) (ln06) {Linear};

\node[draw, fill=Purple!80, rounded corners=1mm, minimum width=3cm, minimum height=6mm, rotate=90] at (7.25,2) (ln07) {BatchNorm};


\node[draw, fill=Emerald, rounded corners=1mm, minimum width=3cm, minimum height=6mm, rotate=90] at (1.50,-2) (ln10) {Linear + GLU};
\node[draw, fill=CadetBlue!80, rounded corners=1mm, minimum width=3cm, minimum height=6mm, rotate=90] at (2.25,-2) (ln11) {Avg. + Tanh};

\node[draw, fill=Emerald, rounded corners=1mm, minimum width=3cm, minimum height=6mm, rotate=90] at (3.50,-2) (ln12) {Linear};
\node[draw, fill=cyan, rounded corners=1mm, minimum width=3cm, minimum height=6mm, rotate=90] at (4.25,-2) (ln13) {GLU + Dropout};
\node[draw, fill=Emerald, rounded corners=1mm, minimum width=3cm, minimum height=6mm, rotate=90] at (5.00,-2) (ln14) {Linear};
\node[draw, fill=cyan, rounded corners=1mm, minimum width=3cm, minimum height=6mm, rotate=90] at (5.75,-2) (ln15) {GLU + Dropout};
\node[draw, fill=Emerald, rounded corners=1mm, minimum width=3cm, minimum height=6mm, rotate=90] at (6.50,-2) (ln16) {Linear};
\node[draw, fill=Purple!80, rounded corners=1mm, minimum width=3cm, minimum height=6mm, rotate=90] at (7.25,-2) (ln17) {BatchNorm};

\node[] at (8.00,-2) (ln18) {$||$};

\draw[->] (input) -- (ln00.north);
\draw[->] (ln00) -- (ln01);
\draw[->] (ln01) -- (ln02);
\draw[->] (ln02) -- (ln03);
\draw[->] (ln03) -- (ln04);
\draw[->] (ln04) -- (ln05);
\draw[->] (ln05) -- (ln06);
\draw[->] (ln06) -- (ln07);

\draw[->] (input) -- (ln10.north);
\draw[->] (ln10) -- (ln11);
\draw[->] (ln11) -- (ln12);
\draw[->] (ln12) -- (ln13);
\draw[->] (ln13) -- (ln14);
\draw[->] (ln14) -- (ln15);
\draw[->] (ln15) -- (ln16);
\draw[->] (ln16) -- (ln17);
\draw[->] (ln17) -- (ln18);

\draw[->] (ln07.south) -- (8.50,2) -- (loss);
\draw[->] (ln18.east)-- (8.50,-2) -- (loss);

\draw[|-|](1,  3.75) -- (2.75, 3.75);
\draw[|-|](3,  3.75) -- (7.75, 3.75);
\draw[|-|](1, -3.75) -- (7.75,-3.75);

\draw[->] (ln00) -- (ln10);
\draw[->] (ln02) -- (ln12);
\draw[->] (ln04) -- (ln14);
\draw[->] (ln06) -- (ln16);

\node[minimum width=6cm, fill=White, inner sep=2pt] at (4.25, 0) () {Update through \textit{Exponential Moving Average}};

\end{tikzpicture}

%% file: figures/knn_embedding.tex
\usetikzlibrary{shapes.misc, positioning}
\begin{tikzpicture}[scale=0.9, transform shape]

\filldraw[draw=black, fill=none, anchor=west] (2.5,4) -- (2.5,8.5) -- (7,8.5) -- (7,4) -- (2.5,4);

\node[circle,fill=black,inner sep=0pt,minimum size=5pt,label=below:{$\vec x_1$}] (a) at (5,5) {};
\node[circle,fill=blue,inner sep=0pt,minimum size=5pt,label=below:{$\vec x_4$}] (a) at (6,5) {};
\node[circle,fill=blue,inner sep=0pt,minimum size=5pt,label=below:{$\vec x_2$}] (a) at (3.4,4.8) {};
\node[circle,fill=blue,inner sep=0pt,minimum size=5pt,label=right:{$\vec x_5$}] (a) at (3,7.1) {};
\node[circle,fill=red,inner sep=0pt,minimum size=5pt,label=left:{$\vec x_3$}] (a) at (6.5,8) {};

\draw [-stealth, dashed, draw=blue](5.15,5) -- (5.9,5);
\draw [-stealth, dashed, draw=blue](4.85,4.95) -- (3.55,4.8);
\draw [-stealth, dashed, draw=blue](4.9,5.1) -- (3.1,7.0);

\node[] at (12.8,4.3) () {Find $k$NN of $\vec x_i \in X$ and create tokens $\vec t_i \in T$};

\node[] at (09.40, 7.90) (xj0) {$\overbrace{\phantom{xxxxx}}^{\textbf{\color{cyan}Global Context}}$};
\node[] at (12.5, 7.55) (xj1) {$\vec t_1 = \big(\textcolor{cyan}{\vec x_1, \vec y_1}, 
\textcolor{orange}{(\vec x_4 - \vec x_1), (\vec y_4 - \vec y_1), (\vec x_2 - \vec x_1), (\vec y_2 - \vec y_1)\ldots} \big)^\T$};
\node[] at (12.5, 7.05) (xj2) {$\vec t_2 = \big(\textcolor{cyan}{\vec x_2, \vec y_2}, 
\textcolor{orange}{(\vec x_1 - \vec x_2), (\vec y_1 - \vec y_2), (\vec x_4 - \vec x_2), (\vec y_4 - \vec y_2)\ldots} \big)^\T$};
\node[] at (12.5, 6.55) (xj3) {$\vec t_3 = \big(\textcolor{cyan}{\vec x_3, \vec y_3}, 
\textcolor{orange}{(\vec x_4 - \vec x_3), (\vec y_4 - \vec y_3), (\vec x_5 - \vec x_3), (\vec y_5 - \vec y_3)\ldots} \big)^\T$};
\node[] at (12.5, 6.05) (xj4) {$\vec t_4 = \big(\textcolor{cyan}{\vec x_4, \vec y_4}, 
\textcolor{orange}{(\vec x_1 - \vec x_4), (\vec y_1 - \vec y_4), (\vec x_2 - \vec x_4), (\vec y_2 - \vec y_4)\ldots} \big)^\T$};
\node[] at (12.54, 5.30) (xj5) {$\vec t_5 = \big(\textcolor{cyan}{\vec x_5, \vec y_5}, \underbrace{
\textcolor{orange}{(\vec x_2 - \vec x_5), (\vec y_2 - \vec y_5), (\vec x_1 - \vec x_5), (\vec y_1 - \vec y_5) \ldots}}_{\textbf{\color{orange}Local Context}} \big)^\T$};

\end{tikzpicture}

%% file: tables/model_hps.tex
\footnotesize
\begin{tabular}{l|rr|rr} 
    \toprule
    \textbf{Parameter} & \multicolumn{2}{c|}{\bf Medium Model} & \multicolumn{2}{c}{\bf Large Model} \\
     Degree of Dim. & $25d$ & $50d$ & $25d$ & $50d$ \\
     \midrule
     $\nu$ & 6 & 6 & 12 & 12 \\
     $k$ & 8 & 8 & 16 & 16 \\
     Num. MHA & 6 & 6 & 6 & 6 \\
     Num. Heads & 4 & 4 & 8 & 8 \\
     $\text{d}_\text{model}$ & 192 & 192 & 384 & 384 \\
     \midrule
     Epochs & 250 & 250 & 250 & 250 \\
     Train Instances & 1\,000\,000 & 1\,000\,000 & 1\,000\,000 & 1\,000\,000 \\
     Batch Size & 1\,264 & 632 & 1\,264 & 632 \\
     Acc. Grad & 1 & 2 & 1 & 2 \\
     \midrule
     $\tau$ & 0.05 & 0.05 & 0.05 & 0.05 \\
     Stride & 1 & 1 & 2 & 2 \\
     EMA Momentum & 0.01 & 0.01 & 0.01 & 0.01 \\
     BN Momentum & 0.1 & 0.1 & 0.1 & 0.1 \\
     \midrule
     Device & \multicolumn{2}{c|}{$3\times$ NVidia Q. RTX 6\,000} & \multicolumn{2}{c}{$3\times$ NVidia RTX A 6\,000} \\
     GPU Memory & \multicolumn{2}{c|}{$3\times$ 24GB} & \multicolumn{2}{c}{$3\times$ 48GB} \\
     Precision & \multicolumn{2}{c|}{Float16 (mixed)} & \multicolumn{2}{c}{BFloat16 (mixed)} \\
     \midrule
     Params. BB. & \multicolumn{2}{c|}{2\,263\,296}  & \multicolumn{2}{c}{9\,189\,888} \\
     Params. Total & \multicolumn{2}{c|}{2\,355\,456}  & \multicolumn{2}{c}{9\,558\,528} \\
     \bottomrule
\end{tabular}

%% file: tables/bbob_hlp.tex
\footnotesize
\begin{tabular}{@{}lrrr@{}}
	\toprule
	{\bf BBOB Function} & {\bf Multim.} & {\bf Global Str.} & {\bf Funnel}\\\midrule
    1: Sphere & none & none & yes \\
    2: Ellipsoidal separable & none & none & yes  \\
    3: Rastrigin separable & high & strong & yes \\
    4: B\"uche-Rastrigin & high & strong & yes \\
    5: Linear Slope & none & none & yes \\ \midrule
    6: Attractive Sector & none & none & yes \\
    7: Step Ellipsoidal & none & none & yes  \\
    8: Rosenbrock & low & none & yes\\
    9: Rosenbrock rotated & low & none & yes\\ \midrule
    10: Ellipsoidal high conditioned & none & none  & yes \\
    11: Discus & none & none & yes\\
    12: Bent Cigar & none & none & yes\\
    13: Sharp Ridge & none & none & yes\\
    14: Different Powers & none & none & yes\\ \midrule
    15: Rastrigin multimodal & high & strong & yes\\
    16: Weierstrass & high & med. & none\\
    17: Schaffer F7 & high & med. & yes\\
    18: Schaffer F7 moderately ill-cond. & high & med. & yes\\
    19: Griewank-Rosenbrock & high & strong & yes\\ \midrule
    20: Schwefel & med. & deceptive & yes\\
    21: Gallagher 101 Peaks & med. & none & none\\
    22: Gallagher 21 Peaks & low & none & none\\
    23: Katsuura & high & none & none\\
    24: Lunacek bi-Rastrigin & high & weak & yes\\\midrule
\end{tabular}

%% file: tables/results_hlp.tex
\resizebox{\textwidth}{!}{
\begin{tabular}{ll||r|rr||rr||rr|rr|rr|rr}
\toprule
 & & ELA$^{*}$ & \multicolumn{2}{c||}{Transformer$^{*}$} & AE-32$^{**}$ & VAE-32$^{**}$ & \multicolumn{2}{c|}{Large (25$d$)} & \multicolumn{2}{c|}{Large (50$d$)} & \multicolumn{2}{c|}{Medium (25$d$)} & \multicolumn{2}{c}{Medium (50$d$)} \\
 High-Level & Dim. & ($50d$) & p100 & p500 & ($50d$) & ($50d$) & RF & SVM & RF & SVM & RF & SVM & RF & SVM \\
\midrule[1pt]
Multi- & 2 & \textbf{0.997} & 0.991 & \textbf{0.997} & 0.849 & 0.856 & 0.854 & 0.845 & 0.895 & 0.904 & 0.896 & 0.905 & 0.938 & 0.939 \\
modality & 3 & \textbf{0.997} & 0.988 & 0.994 & -/- & -/- & 0.869 & 0.877 & 0.946 & 0.953 & 0.958 & 0.950 & 0.974 & 0.973 \\
 & 5 & \textbf{0.999} & 0.991 & \textbf{0.999} & 0.903 & 0.889 & 0.922 & 0.942 & 0.945 & 0.953 & 0.959 & 0.966 & 0.961 & 0.966 \\
 & 10 & \textbf{1.000} & 0.974 & 0.991 & 0.813 & 0.838 & 0.920 & 0.915 & 0.941 & 0.949 & -/- & -/- & -/- & -/- \\
\cmidrule{2-15}
 & all & \textbf{0.998} & 0.986 & 0.995 & -/- & -/- & 0.893 & 0.896 & 0.932 & 0.940 & 0.938 & 0.940 & 0.958 & 0.959 \\
\midrule[1pt]
Global- & 2 & 0.997 & 0.991 & \textbf{0.998} & 0.904 & 0.889 & 0.857 & 0.849 & 0.921 & 0.941 & 0.935 & 0.946 & 0.953 & 0.956 \\
Structure & 3 & \textbf{0.996} & 0.986 & 0.994 & -/- & -/- & 0.890 & 0.909 & 0.954 & 0.968 & 0.960 & 0.961 & 0.976 & 0.977 \\
& 5 & \textbf{0.998} & 0.978 & 0.995 & 0.828 & 0.793 & 0.926 & 0.936 & 0.948 & 0.940 & 0.954 & 0.957 & 0.957 & 0.957 \\
& 10 & \textbf{0.999} & 0.963 & 0.984 & 0.737 & 0.745 & 0.898 & 0.887 & 0.933 & 0.953 & -/- & -/- & -/- & -/- \\
\cmidrule{2-15}
 & all & \textbf{0.998} & 0.980 & 0.993 & -/- & -/- & 0.894 & 0.895 & 0.939 & 0.950 & 0.950 & 0.955 & 0.962 & 0.963 \\
\midrule[1pt]
Funnel- & 2 & 0.998 & 0.999 & \textbf{1.000} & 0.974 & 0.978 & 0.986 & 0.980 & 0.994 & 0.994 & 0.988 & 0.991 & 0.994 & 0.991 \\
Structure & 3 & \textbf{1.000} & \textbf{1.000} & \textbf{1.000} & -/- & -/- & \textbf{1.000} & 0.994 & \textbf{1.000} & \textbf{1.000} & \textbf{1.000} & 0.997 & \textbf{1.000} & \textbf{1.000} \\
& 5 & \textbf{1.000} & \textbf{1.000} & \textbf{1.000} & \textbf{1.000} & 0.991 & \textbf{1.000} & \textbf{1.000} & \textbf{1.000} & \textbf{1.000} & \textbf{1.000} & \textbf{1.000} & \textbf{1.000} & \textbf{1.000} \\
& 10 & \textbf{1.000} & 0.999 & \textbf{1.000} & 0.993 & 0.993 & \textbf{1.000} & \textbf{1.000} & \textbf{1.000} & \textbf{1.000} & -/- & -/- & -/- & -/- \\
\cmidrule{2-15}
 & all & \textbf{1.000} & \textbf{1.000} & \textbf{1.000} & -/- & -/- & 0.996 & 0.994 & 0.999 & 0.999 & 0.996 & 0.996 & 0.998 & 0.997 \\
\bottomrule
\end{tabular}}

%% file: tables/results_so_as.tex
\resizebox{\textwidth}{!}{
\begin{tabular}{ll||r|rr|rr||rr|rr|rr|rr}
\toprule
& & & \multicolumn{2}{c|}{ELA ($50d$)$^{*}$} & \multicolumn{2}{c||}{Transformer$^{*}$} & \multicolumn{2}{c|}{Large (25$d$)} & \multicolumn{2}{c|}{Large (50$d$)} & \multicolumn{2}{c|}{Medium (25$d$)} & \multicolumn{2}{c}{Medium (50$d$)} \\
D & F. Group & SBS$^{*}$ & RF & MLP & p100 & p500 & $k$NN & RF & $k$NN & RF & $k$NN & RF & $k$NN & RF \\
\midrule[1pt]
2 & 1 & \textbf{3.71} & 10.41 & 10.59 & 23.95 & 61.15 & 9.24 & 10.30 & 14.26 & 13.87 & 17.57 & 7.57 & 16.38 & 14.77 \\
 & 2 & 5.80 & 8.51 & 3.72 & 3.54 & 11.04 & \textbf{2.65} & 2.87 & 3.49 & 4.25 & 2.69 & 3.37 & 5.70 & 4.08 \\
 & 3 & 6.29 & 1473.16 & 4.72 & 4.02 & 16.11 & 3.52 & \textbf{3.12} & 5.32 & 4.80 & 3.91 & 4.40 & 5.33 & 4.65 \\
 & 4 & 25.34 & 3.89 & 9.25 & 8.90 & 12.21 & 6.45 & 9.76 & 5.64 & 6.76 & 5.73 & 5.99 & \textbf{3.48} & 4.29 \\
 & 5 & 44.95 & 148.39 & \textbf{3.32} & 4.33 & 6.18 & 3.69 & 5.06 & 3.81 & 7.77 & 3.79 & 8.08 & 3.45 & 14.58 \\
\cmidrule{2-15}
 & all & 17.69 & 342.22 & 6.43 & 9.17 & 21.77 & \textbf{5.21} & 6.36 & 6.63 & 7.62 & 6.91 & 5.99 & 6.92 & 8.66 \\
\midrule[1pt]
3 & 1 & 356.10 & 1480.68 & 11.87 & 13.32 & 39.24 & 19.94 & 66.76 & 15.29 & 15.54 & \textbf{11.72} & 53.23 & 15.32 & 16.13 \\
 & 2 & 4.46 & 8.33 & 3.50 & 2.85 & 6.65 & 2.73 & 3.02 & 3.74 & 3.75 & \textbf{2.67} & 2.87 & 3.42 & 3.62 \\
 & 3 & 4.98 & 7.07 & 3.82 & 2.72 & 9.58 & 2.69 & 3.48 & 3.72 & 4.81 & \textbf{2.59} & 3.53 & 3.85 & 4.02 \\
 & 4 & \textbf{2.63} & 441.96 & 5.06 & 11.49 & 11.87 & 5.15 & 4.63 & 5.20 & 5.00 & 5.36 & 4.83 & 5.12 & 4.20 \\
 & 5 & 66.81 & \textbf{1.22} & 2.54 & 2.46 & 3.04 & 30.75 & 12.02 & 4.21 & 25.80 & 2.70 & 4.99 & 7.24 & 6.85 \\
\cmidrule{2-15}
 & all & 90.43 & 403.67 & 5.44 & 6.72 & 14.39 & 12.65 & 18.61 & 6.54 & 11.28 & \textbf{5.10} & 14.35 & 7.14 & 7.10 \\
\midrule[1pt]
5 & 1 & 11.99 & 14.14 & \textbf{11.97} & 16.27 & 33.39 & 17.70 & 17.31 & 22.88 & 22.81 & 17.50 & 17.85 & 24.18 & 24.01 \\
 & 2 & 3.90 & 369.26 & 2.62 & 4.25 & 5.66 & \textbf{2.40} & 4.27 & 3.03 & 3.89 & 2.48 & 2.45 & 3.59 & 3.39 \\
 & 3 & 4.21 & 150.44 & 3.97 & 5.21 & 9.23 & 3.70 & 4.87 & 4.63 & 6.29 & 3.76 & \textbf{3.69} & 4.33 & 4.58 \\
 & 4 & 4.29 & 1470.28 & 6.81 & 4.33 & 4.47 & \textbf{1.87} & 4.07 & 1.90 & 4.24 & 3.95 & 3.51 & 2.25 & 2.44 \\
 & 5 & 7.67 & 1.13 & 1.83 & 7.72 & 7.93 & \textbf{1.08} & 4.73 & 1.15 & 3.71 & 1.72 & 1.43 & 1.57 & 1.96 \\
\cmidrule{2-15}
 & all & 6.52 & 402.38 & 5.56 & 7.69 & 12.41 & \textbf{5.47} & 7.17 & 6.87 & 8.37 & 6.02 & 5.93 & 7.33 & 7.44 \\
\midrule[1pt]
10 & 1 & \textbf{2.74} & 14.64 & 15.27 & 5.46 & 16.34 & 9.54 & 9.53 & 16.45 & 16.28 & -/- & -/- & -/- & -/- \\
 & 2 & 2.16 & \textbf{1.62} & 1.76 & 2.27 & 2.71 & 2.40 & 2.43 & 2.64 & 2.71 & -/- & -/- & -/- & -/- \\
 & 3 & \textbf{2.76} & 2.87 & 4.35 & 3.10 & 4.48 & 3.54 & 3.62 & 4.52 & 4.15 & -/- & -/- & -/- & -/- \\
 & 4 & 2.02 & 442.01 & 1.96 & 2.03 & 2.09 & 2.06 & 2.05 & \textbf{1.91} & 2.09 & -/- & -/- & -/- & -/- \\
 & 5 & 23.64 & 148.01 & 3.25 & 23.66 & 23.74 & \textbf{1.73} & 11.33 & 1.80 & 12.09 & -/- & -/- & -/- & -/- \\
\cmidrule{2-15}
 & all & 6.85 & 126.84 & 5.46 & 7.51 & 10.17 & \textbf{3.91} & 5.93 & 5.58 & 7.66 & -/- & -/- & -/- & -/- \\
\midrule[1pt]
all & 1 & 93.63 & 379.97 & \textbf{12.43} & 14.75 & 37.53 & 14.10 & 25.97 & 17.22 & 17.13 & 15.60 & 26.22 & 18.62 & 18.30 \\
 & 2 & 4.08 & 96.93 & 2.90 & 3.23 & 6.52 & \textbf{2.54} & 3.15 & 3.23 & 3.65 & 2.62 & 2.90 & 4.24 & 3.69 \\
 & 3 & 4.56 & 408.38 & 4.21 & 3.76 & 9.85 & \textbf{3.36} & 3.77 & 4.55 & 5.01 & 3.42 & 3.88 & 4.50 & 4.42 \\
 & 4 & 8.57 & 589.54 & 5.77 & 6.69 & 7.66 & 3.88 & 5.13 & 3.66 & 4.52 & 5.01 & 4.78 & \textbf{3.62} & 3.64 \\
 & 5 & 35.77 & 74.69 & 2.74 & 9.54 & 10.22 & 9.31 & 8.29 & 2.74 & 12.34 & \textbf{2.73} & 4.83 & 4.09 & 7.80 \\
\cmidrule{2-15}
 & all & 30.37 & 318.78 & \textbf{5.72} & 7.78 & 14.68 & 6.81 & 9.52 & 6.41 & 8.73 & 6.01 & 8.76 & 7.13 & 7.73 \\
\bottomrule
\end{tabular}}

%% file: tables/results_mo_as.tex
\begin{tabular}{l||rr|rr|rr|rr}
\toprule
 & \multicolumn{2}{c|}{Large (25$d$)} & \multicolumn{2}{c|}{Large (50$d$)} & \multicolumn{2}{c|}{Medium (25$d$)} & \multicolumn{2}{c}{Medium (50$d$)} \\
Group & $k$NN & RF & $k$NN & RF & $k$NN & RF & $k$NN & RF \\
\midrule[1pt]
BiBBOB & \textbf{1.000} & \textbf{1.000} & \textbf{1.000} & \textbf{1.000} & \textbf{1.000} & \textbf{1.000} & \textbf{1.000} & \textbf{1.000} \\
DTL & 0.914 & 0.457 & \textbf{1.000} & 0.651 & \textbf{1.000} & 0.651 & \textbf{1.000} & 0.822 \\
MMF & 0.892 & 0.463 & 0.931 & 0.609 & 0.929 & 0.764 & \textbf{1.000} & 0.933 \\
ZDT & -0.079 & -0.087 & 0.041 & 0.209 & 0.688 & 0.185 & 0.616 & \textbf{0.745} \\
\midrule
all & 0.682 & 0.458 & 0.743 & 0.617 & \textbf{0.904} & 0.650 & \textbf{0.904} & 0.875\\
\bottomrule
\end{tabular}